\documentclass[english,twocolumn,transaction]{IEEEtran}
\usepackage{amsmath,amsfonts}

\usepackage{amsthm}
\usepackage{amssymb}
\usepackage{algorithmic}
\usepackage{algorithm}
\usepackage{array, makecell}

\usepackage{float}
\usepackage{setspace} 
\expandafter\def\csname ver@subfig.sty\endcsname{}

\usepackage{svg}

\usepackage{amsfonts}
\usepackage{amssymb}
\usepackage{amsthm}

\usepackage{algorithm}
\usepackage{algorithmic}

\usepackage{stfloats}
\usepackage{subcaption} 
\usepackage{graphicx}
\usepackage{caption}

\usepackage{textcomp}
\usepackage{color}
\usepackage{stfloats}
\usepackage{url}
\usepackage{verbatim}

\usepackage{graphicx}
\usepackage{cite}
\usepackage{booktabs}
\usepackage{hyperref}

\usepackage{bm}
\usepackage{multirow} 
\usepackage{tabularray}

\usepackage{float} 
\usepackage{xcolor}
\begin{document}
	\captionsetup[figure]{font={small}, name={Fig.}, labelsep=period}
	
	\title{Multi-Task Semantic Communication With Graph Attention-Based Feature Correlation Extraction}
	\author{Xi Yu, Tiejun~Lv, \IEEEmembership{Senior Member,~IEEE}, Weicai Li,~\IEEEmembership{Graduate Student Member,~IEEE}, Wei Ni,~\IEEEmembership{Fellow,~IEEE}, \\
		Dusit Niyato,~\IEEEmembership{Fellow,~IEEE},  
		and Ekram Hossain,~\IEEEmembership{Fellow,~IEEE}
        \thanks{Manuscript received April 14, 2024; revised October 27; accepted December 31, 2024. This paper
was supported in part by the National Natural Science Foundation of China
under No. 62271068, and the Beijing Natural Science Foundation under Grant No.
L222046. (\emph{corresponding author: Tiejun Lv.)}}
		\thanks{X. Yu, T. Lv and W. Li are with the School of Information and Communication Engineering, Beijing University of Posts and Telecommunications (BUPT), Beijing 100876, China (e-mail: \{yusy, lvtiejun, liweicai\}@bupt.edu.cn).}
		\thanks{W.~Ni is with the School of Electrical and Data Engineering, University of Technology Sydney, Sydney, NSW 2007, Australia (e-mail: wei.ni@uts.edu.au).}
		\thanks{D. Niyato is with 
  Nanyang Technological University, Singapore 639798 (e-mail: dniyato@ntu.edu.sg).}
		\thanks{E. Hossain is with the University of Manitoba, Canada (e-mail: Ekram.Hossain@umanitoba.ca).}
	}	
	\maketitle
	
	\begin{abstract}
		Multi-task semantic communication can serve multiple learning tasks using a shared encoder model.
		Existing models have overlooked the intricate relationships between features extracted during an encoding process of tasks.
		This paper presents a new graph attention inter-block (GAI) module to the encoder/transmitter of a multi-task semantic communication system, which enriches the features for multiple tasks by embedding the intermediate outputs of encoding in the features, compared to the existing techniques.
		The key idea is that we interpret the outputs of the intermediate feature extraction blocks of the encoder as the nodes of a graph to capture the correlations of the intermediate features. 
		Another important aspect is that we refine the node representation using a graph attention mechanism to extract the correlations and a multi-layer perceptron network to associate the node representations with different tasks. 
		Consequently, the intermediate features are weighted and embedded into the features transmitted for executing multiple tasks at the receiver.
		Experiments demonstrate that the proposed model surpasses the most competitive and publicly available models by 11.4\% on the CityScapes 2Task dataset and outperforms the established state-of-the-art by 3.97\% on the NYU V2 3Task dataset, respectively, when the bandwidth ratio of the communication channel (i.e., compression level for transmission over the channel) is as constrained as \(\frac{1}{12}\).

	\end{abstract}
	
	\begin{IEEEkeywords}
		Semantic communication, multi-task learning, graph attention
	\end{IEEEkeywords}
	
	\section{Introduction}
	\label{Introduction}
	
In data-intensive and real-time applications like the Internet of Things (IoT), challenges arise from not only the sheer volume of data but also many complex tasks, e.g., localization, identification, image segmentation, and tracking. 
Transmitting the raw data can strain bandwidth, particularly in bandwidth-limited (e.g., wireless) systems. 
On the other hand, many tasks involve overlapping data and exhibit correlations~\cite{standley2020tasks}.
In navigation and object detection applications, tasks, e.g., semantic segmentation 
and depth estimation, often operate on the same image data~\cite{chen2018}. 
Semantic communication has emerged as a task-oriented approach that transmits only relevant semantics \cite{shi,bey,tow,com,edgeandsc}. 
By intelligently encoding and transmitting task-specific semantics, it reduces data volume and simplifies processing at the receivers. 
Typically focusing on a single task~\cite{dee, xie,ane,tas,lea,wuq}, semantic communication is increasingly considered to accomplish multiple tasks at one go.
 
As a new paradigm, multi-task semantic communications serve multiple tasks using a unified model and
joint training~\cite{zhang2024unified,aun}. 
A multi-task semantic communication system involves a single encoder and multiple decoders. Each decoder is associated with a specific task and can be deployed 
at different receivers.
The encoder is responsible for generating feature representations for all tasks, while the decoders receive their respective feature representations and execute their tasks.
Multi-task semantic communication is suitable for scenarios where the transmitter (e.g., unmanned aerial vehicles or satellites) can only afford limited hardware and computational complexity, but needs to transmit semantic features for multiple tasks, such as object detection and image segmentation. 
 
A key challenge faced by multi-task semantic communication is an effective design of the encoder architecture, especially under constrained channel conditions with limited bandwidths, so that the feature representations required for different tasks at different decoders can be generated effectively and simultaneously at the encoder.
Existing studies have focused primarily on extracting features for tasks in isolation, 
and they have overlooked the potential benefits of utilizing the feature correlations between tasks. 
There is an opportunity to exploit the correlations and associated features to 
enrich the feature representations to be encoded and transmitted.

Progress has been made on multi-task semantic communication~\cite{zhang2024unified,aun,ima,sem}. 
For instance, Tong et al.~\cite{ima} proposed a task-oriented semantic communication model with an emphasis on reconstruction rate-distortion for image reconstruction and detection. 
Shen et al. \cite{amu} developed a text multi-task semantic communication system based on the bidirectional encoder representations from transformers (BERT)~\cite{bert}, where text semantics were extracted at the transmitter for text classification and regression tasks.
The Semantic-based Multi-level Feature Extraction Model (SMFEM) \cite{sem} was designed to capture multi-layer features with several encoders to accomplish different tasks. Recently, the authors of \cite{aun} put forth a unified semantic communication system (U-DeepSC) using domain adaptation to tailor features for different tasks to reduce storage and training redundancy. The study in~\cite{zhang2024unified} further advanced U-DeepSC by designing a lightweight feature selection module, adapting to varying channel conditions.
	
	\subsection{Related Works}
 Task-oriented semantic communication systems 
 prioritize the precise transfer of task information over the exact replication of the entire data or signal content \cite{ama}. Initial research focused on single-task semantic communication, e.g., images, voice, and text, with a dedicated pair of encoder and decoder designed for each task. For instance, Bourtsoulatze et al. \cite{dee} introduced Deep Joint Source-Channel Coding (DJSCC) for image transmission in wireless channels, utilizing CNNs to extract features. Xie et al.~\cite{xie} proposed DeepSC, a semantic communication architecture using Transformers to extract text semantics. Considering speech, Shi et al. \cite{ane} proposed a comprehension-before-transfer framework with high semantic fidelity and verified its effectiveness in audio transmission. Moreover, Wu et al. \cite{wuq} presented a Grad-CAM-based method for semantic transfer between tasks.

Multi-task semantic communication has been increasingly considered to improve the utilization of computational and storage resources. For instance, the authors of~\cite{sem} presented a coarse-to-fine architecture for image feature extraction and compression, enhancing reconstruction performance and adaptable to various tasks by modifying the decoding layers. In \cite{ima}, the TOSC-SR scheme was developed by extending rate-distortion optimization theory to enhance the performance of multiple tasks while ensuring image reconstruction quality. The authors of \cite{amu} built a multi-task semantic communication system using BERT~\cite{bert}, which can handle text-related tasks by extracting and transmitting semantics from text at the transmitter. The authors of \cite{sca} introduced a scalable multi-task semantic communication system that dynamically adjusts the encoding rate based on the importance ranking of features for different tasks.
The authors of \cite{aun} proposed a unified deep learning-based semantic communication system, which employs domain adaptation techniques at the transmitter to tailor the shared and private features for different tasks and adjusts decoding efficiency at the receiver. U-DeepSC \cite{zhang2024unified} was further developed utilizing a vector-based dynamic scheme and a lightweight feature selection module to adapt to different tasks and channel conditions. It was trained for a single task within a shared encoder, rather than dealing with multiple tasks simultaneously.

In the context of computer vision (CV) and natural language processing (NLP), multi-task learning has been developed, which can be classified into two categories. 
The first category focuses on efficient feature sharing across tasks, e.g., Cross-Stitch \cite{cro}, which linearly combines features from different task layers, and Sluice Network~\cite{lat}, which uses a gating strategy for flexible feature sharing. 
Other examples include MTAN \cite{endt}, which 
links task-specific attention networks to a shared backbone, and NDDR-CNN \cite{nddr}, which introduces an asymmetric domain adaptation layer to manage object detection and semantic segmentation tasks concurrently.
The second category 
comprises models emphasizing dynamic network selection. For instance, DEN~\cite{ahn} integrates a selector to designate subnetworks based on input data, and AdaShare~\cite{ada} selects network blocks for tasks via a task-specific policy. In light of this, Dynashare \cite{rahimian2023dynashare} captures input characteristics for encoding path selection.

These existing models~\cite{sem,ima,amu,zhang2024unified,cro,lat,endt,nddr,ahn,ada,rahimian2023dynashare} still face constraints in deploying flexible and effective encoding strategies tailored to different tasks. Accurately extracting semantic features needed for the execution of multiple tasks is of paramount importance, particularly when operating under restricted communication bandwidth. Moreover, none of the existing models has attempted to capture the correlations among features extracted in the encoders.

	\subsection{Contribution}
	
	This paper presents a new graph attention inter-block (GAI) module to the encoder/transmitter of multi-task semantic communication to enrich features captured and delivered over constrained communication channels, thereby enhancing the completion of multiple tasks simultaneously. 
	The key novelty is that we interpret the output of each feature extraction block of the encoder as a node, and then build a graph to effectively reveal the correlations among the extracted intermediate features.
	Another important advancement is that we update the node representation using a graph attention mechanism to capture the correlations among the intermediate features.
	The connections among the intermediate features and the tasks are adaptively adjusted to meet specific task requirements by generating task-node weights through a multi-layer perceptual network. 
	A task-node weight is a tensor with the same size as the output channel, accounting for the contribution of the corresponding feature to a task. 
	
	The contributions of this paper are summarized as follows:
	\begin{itemize}
		\item 
		We develop a new GAI module to capture the correlation among the intermediate features extracted from different feature extraction blocks of the encoder of a multi-task semantic communication system to improve the utilization of each encoding block and enrich the representations of the features delivered.
		\item 
		We interpret the intermediate features extracted at different feature extraction blocks of the encoder as nodes on a graph. The intermediate features are unified by a Feature Transformation Layer into a standardized node representation. Their correlations are captured iteratively and enhanced by employing a Graph Attention Layer in the GAI module. 
		\item 
		We design a Relation Mapping Layer in the GAI module to generate task-node weights by feeding the node representations generated by the Feature Transformation Layer into task-specific multi-layer perceptual networks to satisfy task-specific requirements. 
	\end{itemize}  
	
	Extensive experiments demonstrate that when the communication channel is efficient (e.g., with a bandwidth ratio of 1.3), the proposed GAI notably improves average accuracy across all tasks. It surpasses the leading baseline by 2.71\% on the CityScapes 2Task dataset; outperforms the respective leading baselines by at least 0.87\% and 0.80\% on the NYU v2 dataset for two-task and three-task semantic communications, respectively; by 9.47\% on the TaskonomyTiny 5Task dataset; and by 1.2\% and 5.19\% on the Oxford-IIIT Pet 
    and MVSA datasets,
    respectively.
	When the bandwidth ratio is reduced to as small as \(\frac{1}{12}\), the GAI outperforms the baselines by 11.4\% and 3.97\% on the CityScapes 2Task and NYU V2 3Task datasets, respectively.
	The performance improvement becomes more pronounced as the SNR decreases.

	
	
	The rest of this paper is structured as follows: Section II outlines the system model. Section III delineates the proposed GAI model for the effective extraction of features and their correlations for multi-task semantic communication. Section IV evaluates the performance of the proposed GAI across various datasets, bandwidth ratios, and SNRs, including visualization of how different block coding relates to specific tasks. Finally, Section V concludes the paper with a summary of our findings and conclusions.
	
	\begin{table}[htbp]
		\caption{Notation and definitions}
		\centering
		\footnotesize
		\begin{tabular}{cl}
			\toprule
			Notation & Description \\
			\midrule
			\( \mathbf{\varphi} \) & Encoding parameters\\
			\(\mathbf{\theta}\) & Decoding parameters \\
			\( T \) & Number of tasks\\
			\( \mathbf{x} \) & Input data \\
			\( \mathbf{\hat{y}}\) & Decoded data  (dimension varies with task) \\
			\( \mathbf{z} \) & Transmitted data\\
			\( \mathbf{\hat{z}} \) & Received data\\
			\( \mathbf{\eta}\) & Channel noise\\
			\( \mathbf{\sigma}^2\) & Noise variance \\
			\( \alpha\) & Width and height reduction coefficient  \\
			\( R\) & Bandwidth ratio\\
			\( \mathbf{F_i} \) & The output of the encoded feature of the $i$-th block. \\
			\( \mathbf{K_i} \) & The $i$-th block feature after InterpolateLayer. \\
			\( \mathbf{V}_i^M\) & The $i$-th node encoded representation after $M$ updates. \\
			\( \mathbf{U}\) & Transformation matrix for computing attention coefficients. \\
			\( \mathbf{P}\) &  Transformation matrix for updating node representations. \\
			\( a_{i,j}\) & Attention value between nodes $i$ and $j$. \\
			\( \mathbf{e}_{i,t}\) & Task-node relevance of node $i$ to task $t$. \\
			\bottomrule
		\end{tabular}
		\label{table:1}
	\end{table}
	
	\section{System Structure, Task Model, and Use Case}
	\label{sec:System Architecture, Task Model, and Application Scenario}
	
	\begin{figure}
		\centering
		\includegraphics[width=0.475\textwidth]{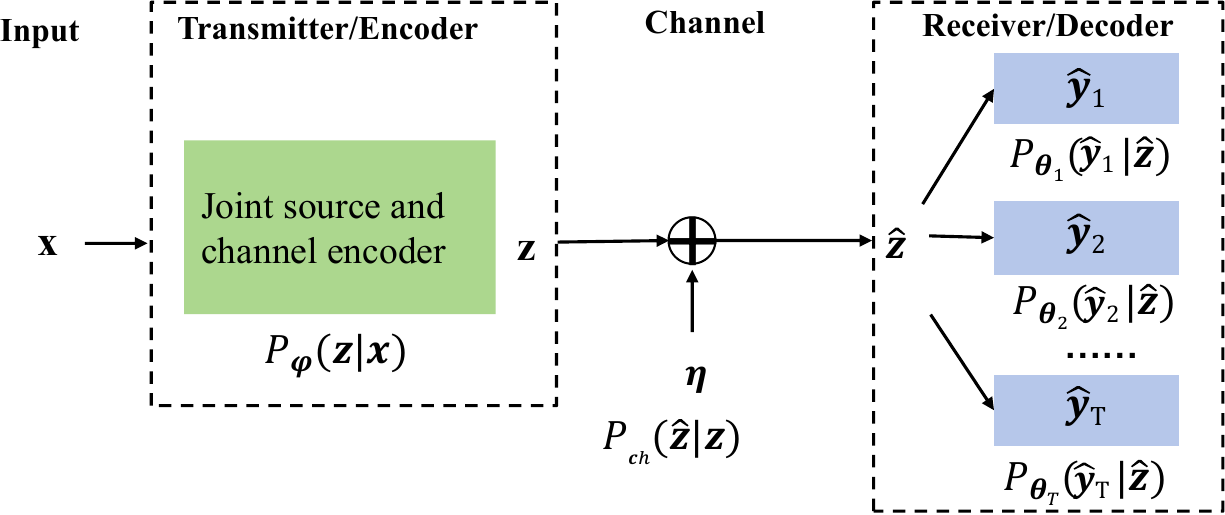}
		\caption{The proposed multi-task semantic communication system model, where the encoder extracts features, which are transmitted through the channel to the decoders for multiple simultaneous tasks.}
		\label{fig_1}	
	\end{figure}
	
	\begin{figure*}[!ht]
		\centering
		\includegraphics[width=0.65\linewidth]{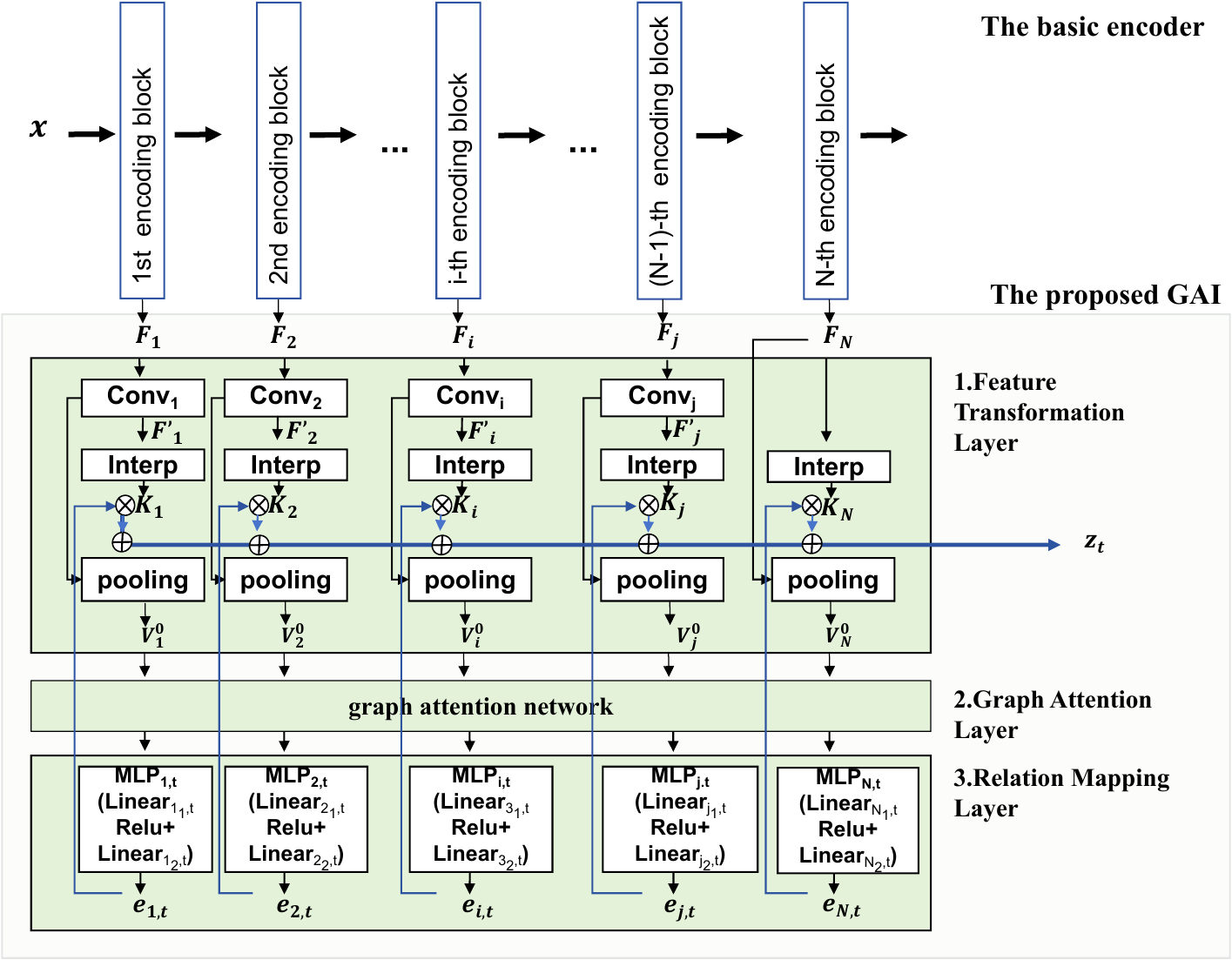}
		\caption{   
			\small 
The proposed GAI module, which involves generating encoded data $\mathbf{z}_t$ for task $t$ within the GAI module. This process begins with inputs from the output of each basic encoder block. The procedure comprises (i) a Feature Transformation Layer, (ii) a Graph Attention Layer, and (iii) a Relation Mapping Layer. The Feature Transformation Layer unifies each block's feature size, and the Graph Attention Layer updates the node representation to interact with the other nodes. The Relation Mapping Layer generates task-node weights specific to each task, denoted as $\mathbf{e}_{i,t}$. $\mathbf{e}_{i,t}$ is merged with $\mathbf{K}_i$ to obtain the task-specific transmission representation $\mathbf{z}_t$ for each respective task.}
		\label{fig:first_figure}
	\end{figure*}
	
	\begin{figure}[t]
		\centering
		\includegraphics[width=0.5\linewidth]{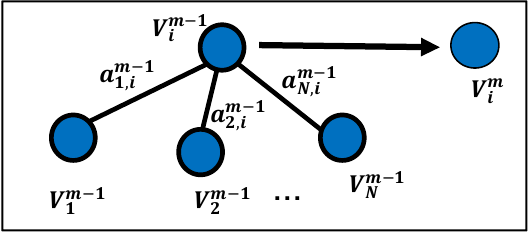}
		\caption{
			\small
			Graph Attention Layer: This diagram illustrates the iterative process of updating representation for the $i$-th node, specifically focusing on the $m$-th update iteration. Each node undergoes a total of $M$ iterations for information updating.
		}
		\label{fig:second_figure}
	\end{figure}
	
	The considered multi-task semantic communication system trains an end-to-end multi-task learning model over a bandwidth-limited wireless channel.
	The system consists of a transmitter with an encoder and a receiver with multiple decoders at both ends of the channel; see Fig. \ref{fig_1}.
	Unlike traditional communication systems with the primary objective of preserving data fidelity by performing the source and channel coding separately, the encoder here extracts and encodes task-relevant features from the input data. 
	Upon receiving the encoded data, the decoders 
 process the encoded features directly to accomplish multiple tasks. 

	\subsection{Task Model}
	\label{subsec:Task Model}
	Given the input (image) data \(\mathbf{x} \in \mathbb{R}^{C_{\text{in}} \times W_{\text{in}} \times H_{\text{in}}}\) to the encoder, multiple complex tasks need to be completed at the decoder in the receiver. 
	\(C_{\text{in}}\), \(W_{\text{in}}\), and \(H_{\text{in}}\) represent the channel number, width, and height of the input data, respectively. 
	The decoding output of the $t$-th task is \(\hat{\mathbf{y}}_t\), \(t =1, \cdots, T\), with \(T\) being the number of tasks.
	In the case of semantic segmentation, $\hat{\mathbf{y}}_t$ indicates the decoded pixel category.
	In the case of depth prediction, $\hat{\mathbf{y}}_t$ denotes the model's decoded depth value.
	In the case of surface normal estimation, $\hat{\mathbf{y}}_t$ provides the estimated vectors of the surface normal for each pixel.
	
	\subsection{Encoder and Decoder}
	\label{subsec:Encoder and Decoder}
 
 Our primary focus is on the encoder's design on the transmitter side. We propose to enhance the feature maps to be transmitted by capturing their correlations so that the decoders at the receiver side can achieve improved task performances without modifications to the decoders.
	\subsubsection{Encoder}
	\label{subsubsec:Encoder}
	The encoder performs joint source-channel coding on the input data $\mathbf{x}\in \mathbb{R}^{C_{\text{in}} \times W_{\text{in}} \times H_{\text{in}}}$, which extracts and encodes the features required for multiple tasks at the decoder.
	The output of the encoder is denoted by \(\mathbf{z} \in \mathbb{R}^{C_{\text{out}} \times W_{\text{out}} \times H_{\text{out}}}\), 
	where $C_{\rm out}$, $W_{\rm out}$, and $H_{\rm out}$ are the channel number, width, and height of the output, respectively. 
	
	The encoding process can be characterized as a conditional distribution, as given by 
	$$p_{\boldsymbol{\varphi}} (\mathbf{z}|\mathbf{x}),$$ 
	where \(\boldsymbol{\varphi}\) denotes the model parameters of the encoder.
	After encoding, \(\mathbf{z}\) is transmitted through the channel to the receiver.
	Considering the hardware complexity of the transmitter, we employ a shared encoder that employs a ResNet architecture with multiple, sequentially connected residual blocks.

	\subsubsection{Decoder}
	\label{subsubsec:Decoder}
	At the receiver, there are $T$ task-specific joint source-channel decoders, each specialized for a specific task.
	Each decoder, i.e., the $t$-th decoder, is responsible for decoding the received signal, denoted by $\hat{\mathbf{z}}$, based on the requirements of the corresponding task $t$, generating the result \(\mathbf{\hat{y}}_t\). 
	
	Every decoder consists of several convolutional layers, each meticulously designed to handle the specificity of the corresponding task. 
	The following conditional probability distribution can represent the operation of the $t$-th decoder:
	\begin{align}
		p_{\boldsymbol{\theta}_t} (\mathbf{\hat{y}}_t|\hat{\mathbf{z}}),\,t=1,\cdots,T,
	\end{align}
	where the model parameters $\boldsymbol{\theta}_t$ are tailored for task $t$. The decoders operate independently. Their parameters $\boldsymbol{\theta}_t$ are tuned according to the specific requirements of each task~$t$.
	
	\subsection{Channel Model}
	\label{subsec:Channel model}
	The encoder/transmitter and the decoder/receiver are connected by a bandwidth-limited and noisy channel. 
	The input and output of the channel are $\mathbf{z}$ and $\hat{\mathbf{z}}$, respectively. 
	The transmission in the channel process can be characterized by a conditional probability, as given by $$p_{\text{ch}}(\hat{\mathbf{z}}|\mathbf{z}).$$
	Without loss of generality, we use additive white Gaussian processes to model the channel\footnote{The technique developed in this paper can be readily extended to other channel models, such as Rayleigh fading channels.}, i.e., 
	\begin{equation}
		\hat{\mathbf{z}} = \mathbf{z} + \boldsymbol{\eta}, 
	\end{equation}
	where $\boldsymbol{\eta} \in \mathbb{R}^{C_\text{out}\times H_\text{out} \times W_\text{out}}$ collects the receiver noise with each element sampled from $\mathcal{N}(0, \sigma^2)$. Here, $\mathcal{N}(0, \sigma^2)$ denotes the zero-mean Gaussian distribution with variance~$\sigma^2$. 
	
	At the receiver, the SNR of the received signal is given by
	\begin{equation}
		\text{SNR (dB)} = 10 \log \left(P/\sigma^2\right),
	\end{equation}
	where $P$ is the transmit power of the encoded signal $\mathbf{z}$. 
	
	Let \( n \) denote the input data size (in pixels). Then, $n = C_{\text{in}} \times W_{\text{in}} \times H_{\text{in}}$.
	Also, let \( k \) denote the size (in pixels) of the input to the channel. Then, 
	\begin{equation}
		k = C_{\text{out}} \times W_{\text{out}} \times H_{\text{out}}/2,
	\end{equation}
	where the denominator~``2'' accounts for the transmission of the complex latent vectors in the channel.
	As a result of encoding, the dimension of the input feature is transformed as
	\begin{equation}
		\begin{aligned}
			W_{\text{out}} &= \frac{W_{\text{in}}}{\alpha} \text{ and }
			H_{\text{out}} = \frac{H_{\text{in}}}{\alpha},
		\end{aligned}
	\end{equation}
	where $\alpha$ is known as the reduction coefficient for the height and width of the input feature during encoding. $\alpha$  depends on the stride settings in the convolutional operations in the ResNet architecture. 
	As a result, the widths and heights of the features extracted at the residual blocks decrease progressively from the input to the output of the encoder.
	
	With reference to \cite{djscc-f,xu2021wireless,dee}, the bandwidth ratio of the channel, denoted by \( R \), is $R = \frac{k}{n}$, which indicates the level to which the extracted features need to be compressed for transmissions over the channel with limited bandwidth.
	
	\subsection{End-to-End Multi-Task Semantic Communication Model}
	
	The overall probability of accurately encoding and decoding the label $\hat{\mathbf{y}}_t$, $t=1,\cdots, T$, is expressed as 
	\begin{equation}
		p(\mathbf{\hat{y}}_t|\mathbf{x}) = p_{\boldsymbol{\theta}_t} (\mathbf{\hat{y}}_t|\hat{\mathbf{z}}) \cdot p_{\text{{ch}}}(\hat{\mathbf{z}}|\mathbf{z}) \cdot p_{\boldsymbol{\varphi}} (\mathbf{z}|\mathbf{x}), t=1,\cdots, T.
	\end{equation}
	
	\section{Graph Attention Inter-block multi-task Semantic Communication}
	In this section, we delineate the new GAI module to implement the encoder for effective multi-task transmission and execution at the decoders.
	The GAI module interprets the output of each feature extraction residual block of the encoder as a node in a fully-connected graph. 
	Then, the graph attention mechanism is employed to quantify the relationships among the nodes (or the feature extracted by the blocks). Correlations among features extracted at different residual blocks can be captured to enrich the features with an additional dimension reflecting the correlations across the outputs of the residual blocks.
	This differs from traditional methods, which linearly encode features through blocks and often inadvertently overlook the encoded details from the preceding blocks.
	
	The GAI module comprises $(a)$ a Feature Transformation Layer, which applies convolution and interpolation to normalize the sizes of features extracted at different blocks. $(b)$ a Graph Attention Layer, which employs graph attention mechanisms to transform the outputs of the feature extraction blocks into nodes on a graph, and $(c)$ a Relation Mapping Layer, which generates the task-node weights on the graph. These weights are utilized to fine-tune the importance of each feature in line with the specific requirements of each task.
	
	It is noted that the Feature Transformation and the Graph Attention Layers are task-agnostic, enabling generalization across various tasks. The Relation Mapping Layer is customized for each task, enhancing task-specific performance.

	\subsection{Feature Transformation Layer}
	\label{subsec: Feature Transformation Layer}
	
	The ResNet encoder comprises multiple residual blocks. Each block is an independent feature extraction encoder unit. Let \(N\) denote the number of such blocks. Their outputs are~\({\mathbf{F}_1, \cdots, \mathbf{F}_N}\) with \( \mathbf{F}_i\in \mathbb{R}^{C_i \times W_i \times H_i}\), \(i = 1, \cdots, N\) being the output of the \( i \)-th block. The Feature Transformation Layer unifies the latent features generated by these blocks into the initial standardized node representation in an $N$-vertex graph. 
	The vertices of the graph retain the latent features from their corresponding residual blocks and are enhanced through the exchange of information with other nodes. 
	
	The objective of this layer is to convert latent features from various residual blocks into node representations with consistent dimensions. To accomplish this, we design a series of processes involving convolution layer transformation, pooling, and interpolation layer adjustments.
	
	The convolution layer transforms the latent features with different channel sizes into the target output channel size while retaining critical spatial information, as given by
	\begin{equation}
		\text{\textit{ConvLayer}}: [C_i, W_i, H_i] \rightarrow [C_{\text{out}}, W_i, H_i].
		\label{eq:convlayer} 
	\end{equation}	
 
	This convolution layer involves multiple convolution kernels (filters), i.e., Conv$_i$ in Fig.~\ref{fig:first_figure}, to progressively capture the local features of the input $\mathbf{F}_i$. 
These filters scan over the input data, apply the convolution operation at each position, and generate a series of feature maps, denoted by 
$\mathbf{F}'_i\in \mathbb{R}^{C_{out} \times W_i \times H_i}, \forall i$, to retain  semantic features from the input:
		\begin{equation}
			\mathbf{F}'_i = \mathbf{Q}_i * \mathbf{F}_i + \mathbf{b}_i,\,\forall i,
            \label{eq:convlayer2}
		\end{equation}
where \(\mathbf{Q}_i\) is the weights of the convolutional filters of the $i$-th block, 
\(\mathbf{b}_i\) is a bias of the $i$-th block, and 
\(*\) denotes the convolution operation. 
The feature maps, i.e., $\mathbf{F}'_i, \forall i$, capture complex patterns and information, contributing to the subsequent generation of node representations.
	
	The pooling operation further refines the feature dimension of each node, $\mathbf{F}'_i, \forall i$, as given by 
	\begin{equation}
		\text{\textit{Pooling}}: [C_{\text{out}}, W_i, H_i] \rightarrow [C_{\text{out}}].
		\label{eq:pooling}
	\end{equation}
	With the convolution and pooling operations, we ensure the consistency in the output size of each feature extraction block when transforming $\mathbf{F}_i$ through $\mathbf{F}'_i$ to an initial node representation~$\mathbf{V}_{i}^0 \in \mathbb{R}^{C_{\text{out}} \times 1},\,\forall i=1,\cdots,N$.
	
	We also incorporate an interpolation layer to ensure consistent dimension sizes during the final fusion of node representation. The interpolation layer adjusts the heights and widths of the features $\mathbf{F}'_i$ to unify their sizes. This is achieved by utilizing the weighted average of the nearest four-pixel values for resizing. It can preserve the fidelity of the original features while ensuring smooth and accurate size transitions.
	The interpolation process can be written as
	\begin{equation}
		\text{\textit{InterpolateLayer}}: [C_{\text{out}}, W_i, H_i] \rightarrow [C_{\text{out}}, W_{\text{out}}, H_{\text{out}}].
		\label{eq:interpolation} 
	\end{equation}
	We use $\mathbf{K}_i \in \mathbb{R}^{C_{\text{out}} \times W_{\text{out}} \times H_{\text{out}}}$ to denote 
 the interpolated version of $\mathbf{F}'_{i}, \, i=1,\cdots,N$.

	\subsection{Graph Attention Layer}
	\label{subsec: Graph Attention Layer}

The initial node representations $\mathbf{V}_{i}^0,\, i=1,\ldots,N,$ are input into the Graph Attention Layer, where they are iteratively updated by adjusting the relationships between the nodes. 
By leveraging the attention mechanism, the Graph Attention Layer evaluates the importance of neighboring nodes' features correlating each node, allowing the model to refine each node's representation by integrating relevant information from its connections. 
This process results in more accurate and task-specific node representations that can better capture the underlying graph structure.
	
	
	As shown in Fig.~\ref{fig:second_figure}, we employ Graph Attention Network (GAT) to tune the representations of each node. GAT stands out as a sophisticated structure tailored for analyzing interconnected graph data~\cite{gatsurvey}. It assesses the interconnections within a graph by employing an attention mechanism, where the states of neighboring nodes are considered when updating a node's state. 
	Moreover, GAT can capture multiple features to optimize node representations by utilizing multiple attention heads.
	
	In the encoder, correlations exist among the features encoded by different feature-extracting residual blocks. The attention mechanism can dynamically assign importance to each neighbor, enabling the model to prioritize relevant information and thus refine the representations of a node. Leveraging the attention mechanism in GAT, we update the representation of each node and establish relationships with every other node. By further constructing the relationship between each node and the task, the updated node representation comprehensively integrates information from other nodes, resulting in more accurate task-node weights. 
	
	The node representations are updated through an iterative process involving information interactions between the nodes. 
	Assume the node representations are updated for a total of \( M \) times. During the \( m \)-th iteration, with \( m = 1, \cdots, M \), for nodes \( i \) and \( j \), a learnable weight matrix \( \mathbf{U}^m \in \mathbb{R}^{C_{\text{out}} \times C_{\text{out}}} \) and the LeakyReLU activation function are used to compute the attention coefficients, denoted by \( a_{i,j}^m \). The Softmax function, softmax\((\cdot)\), is applied at each node to normalize the attention coefficients across all adjacent nodes, as given by 
	\begin{equation}
		a_{i,j}^m = \text{softmax}_j \left( \text{LeakyReLU} \left( \mathbf{a}^\top [ \mathbf{U}^m \mathbf{V}_i^{m-1} \, || \, \mathbf{U}^m \mathbf{V}_j^{m-1}] \right) \right),
		\label{eq:attention}
	\end{equation}
	where \( \mathbf{a} \in \mathbb{R}^{2C_{\text{out}}} \) is a learnable weight vector, and 
	$\mathbf{V}_i^{m-1} \in \mathbb{R}^{C_{\text{out}} \times 1}$ is the representation of node $i$ after the $(m-1)$-th iteration. 
 $(\cdot)^\top$ stands for transpose. \( (\cdot || \cdot)\) denotes concatenation.
 
 Then, $\mathbf{V}_i^{m} $ is derived by calculating the weighted sum of the representations of node $i$ and its neighboring nodes collected in the set $\mathcal{H}(i)$, as given by
	\begin{equation}
		\mathbf{V}_i^{m} = \text{ReLU} \Big( {\sum}_{j \in \mathcal{H}(i)} (a_{i,j}^m \cdot \mathbf{P}^m \mathbf{V}_{j}^{m-1}) \Big),
		\label{eq:updating}
	\end{equation}
	where $\mathbf{P}^m \in \mathbb{R}^{C_{\text{out}} \times C_{\text{out}}}$ is a learnable weight matrix used for refining node representations.	
	After $M$ updates, the final node representation is obtained, i.e., ${\mathbf{V}_{\text{1}}^M, \cdots, \mathbf{V}_{N}^M} \in \mathbb{R}^{C_{\text{out}} \times 1}$. 

Compared to a simpler attention mechanism~\cite{vaswani2017attention}, the GAT benefits from the additional information explicitly modeled in the graph structure.
In a simple attention network, each node \( i \) has an initial feature vector.
The features are used to compute query,
key,
and value 
vectors for attention score calculation
to capture the relationships based on node features 
without modeling the connections between the nodes.
In contrast, the Graph Attention Layer exploits the graph structure to incorporate additional information during attention score calculation. 
Specifically, GAT uses the adjacency information to dynamically adjust the importance of neighboring nodes.
By concatenating the features of connected nodes and applying a learnable attention vector, GAT captures richer non-linear relationships within the graph. 
The non-linearity introduced by the LeakyReLU function extends the capability of the model to discern complex patterns.
	
	\subsection{Relation Mapping Layer}
	\label{subsec: Relation Mapping Layer}
	
	The Relation Mapping Layer is specifically designed to establish connections among the encoding node representations, i.e., $\mathbf{V}_i^M$, and various tasks, thereby maximizing the utilization of encoding feature from every block to each task.
	The Relation Mapping Layer consists of a multi-layer perceptual network incorporating linear and activation layers. 
	
	To tailor the encoding feature for diverse tasks, the final representation of each node $i$, i.e., $\mathbf{V}_i^{\text{M}}$, 
	is first passed through two linear networks, denoted by $Linear_{i_1,t}$ and $Linear_{i_2,t}$, with a ReLU activation function connecting them, generating task-node weights $\mathbf{e}_{i,t}$ for each task $t$, as given by 
	\begin{equation}
		\mathbf{e}_{i,t} = \text{Linear}_{i_2,t}(\text{ReLU}\left(\text{Linear}_{i_1,t}(\mathbf{V}_i^{\text{M}})\right),
		\label{eq:nodeencoding}
	\end{equation}
	where $\mathbf{e}_{i,t} \in \mathbb{R}^{C_{\text{out}}}$ stands for the weight of node $i$ in regards of task $t$ in the interpolated node representation $\mathbf{K}_i$ to be transmitted, and hence, it has $C_{\text{out}}$ channels. Each task generates corresponding weights at each node, with a total of \(T \times N\) task-node weights, quantifying the importance and contribution of each encoding node in handling specific tasks.
	
	After calculating the task-node weights $\mathbf{e}_{i,t}$, a fusion method is employed so that these weights are combined with the corresponding features $\mathbf{K}_i$ to generate the feature for the execution of each task $t$, as given by 
	\begin{equation}
		\mathbf{z}_t = {\sum}_{i=1}^N (\mathbf{e}_{i,t} \cdot \mathbf{K}_i),\, \forall t = 1,  \cdots, T,
		\label{eq:finaltaskencoding}
	\end{equation}
	where $\mathbf{z}_t \in \mathbb{R}^{C_{\text{out}} \times W_{\text{out}} \times H_{\text{out}}}$ represents the final transmitted encoding feature specifically tailored for task $t$.
	
	\subsection{Objective Function} 
	\label{Objective Function}
	The primary objective of the considered multi-task semantic communication system is to minimize transmission loss across all tasks by optimizing the parameters of the designed encoder, denoted by $\boldsymbol{\varphi}$, and the decoder parameters, denoted by $\boldsymbol{\theta}$.
	The parameter set $\boldsymbol{\varphi}$ includes the parameters of the encoder and the GAI module. 
	The parameter set $\boldsymbol{\theta}$ comprises of parameters $\boldsymbol{\theta}_1, \ldots, \boldsymbol{\theta}_T$ for the \(T\) task-specific decoders. Each decoder consists of three convolutional layers.
	
	The total loss of the multi-task semantic communication framework can be written as the following weighted sum of the losses of the tasks: 
	\begin{equation}
		L\left( \boldsymbol{\theta},\boldsymbol{\varphi}\right) = {\sum}_{i=1}^{T} w_t L_t,
	\end{equation}
	where $L_t$ denotes the loss for task $t$, and  $w_t$ is the corresponding weight. 
	$\left\{w_1,\ldots,w_T\right\}$  are configured to ensure that the loss for each task is of the same magnitude. 
	
	The optimization problem is thus formulated as
	\begin{equation}
		\underset{\boldsymbol{\theta}, \boldsymbol{\varphi}}\min \ L\left( \boldsymbol{\theta},\boldsymbol{\varphi}\right) .
	\end{equation}
	Optimizing these parameters is crucial for reducing the total loss across all tasks, which, in turn, enhances the overall efficacy of the multi-task learning system. For example, the loss functions for different tasks, $L_t,\, t=1,\ldots,T$, can be defined as follows:
	
	\subsubsection{Semantic Segmentation Loss}
	Semantic segmentation involves classifying each pixel in an image into a predefined category. The loss function can be defined as
	\begin{equation}
		L_{seg} = -\mathbf{y}_{gt} \cdot \log(p({\mathbf{\hat{y}}})),
	\end{equation}
	where $\mathbf{y}_{gt}$ is the true label and $p({\mathbf{\hat{y}}})$ is the predicted category probability for a pixel.
	
	\subsubsection{Loss for Depth Prediction, Keypoint Detection, and Edge Detection}
	
	Depth prediction and keypoint detection, and edge detection tasks employ a similar loss function to quantify the difference between predictions and ground-truth values. The unified loss function is given by
	\begin{equation}
		L_{dep/key/edg} = |\hat{y} - y_{gt}|,
	\end{equation}
	where $\hat{y}$ is the predicted value and $y_{gt}$ is the corresponding ground-truth. In the context of depth prediction, $\hat{y}$ and $y_{gt}$ are the predicted and ground-truth depth values, respectively, aiming to estimate the distance of each object in an image from the viewpoint. For keypoint detection, $\hat{y}$ and $y_{gt}$ correspond to the predicted and ground-truth keypoint values, with the goal of identifying points of interest in an image. For edge detection, $\hat{y}$ and $y_{gt}$ are the predicted and ground-truth edge values, respectively. 
	
	\subsubsection{Surface Normal Estimation Loss}
	Surface normal estimation involves calculating the angle of surfaces relative to the viewer. The loss function is given by
	\begin{equation}
		L_{sn} = 1 - \frac{\mathbf{\hat{y}} \cdot \mathbf{y}_{gt}}{|\mathbf{\hat{y}}| \cdot |\mathbf{y}_{gt}|},
	\end{equation}
	where $\mathbf{\hat{y}}$ is the predicted normal and $\mathbf{y}_{gt}$ is the true normal, both of which are normalized to unit length. 
	
	\subsection{Workflow and Complexity Analysis of GAI}
 \label{sec:Computational Complexity of Algorithm}
	\textbf{Algorithm 1} summarizes the proposed GAI algorithm, which commences with feeding data into the encoder; see Section~\ref{subsubsec:Encoder}. Each encoding block in the encoder produces an encoding feature. 
	The features are fed to the Feature Transformation Layer, which adjusts the channel from each block to a unified size and refines these adjusted features into initial node representations; see Section~\ref{subsec: Feature Transformation Layer}. The spatial dimensions of features are unified by interpolation.
	Next, the Graph Attention Layer updates the node representations using a GAT, where the attention coefficients are computed between node pairs to obtain the correlations; see Section~\ref{subsec: Graph Attention Layer}. 
	Then, the updated node representations are processed to generate task-node weights in the Relation Mapping Layer, where the weights are combined with block features to produce features to be transmitted; see Section~\ref{subsec: Relation Mapping Layer}. 
	The receiver decodes the received encoding features to execute specific tasks, as described in Section \ref{subsubsec:Decoder}.

  
	To analyze the computational complexity of \textbf{Algorithm~\ref{algorithm:1}}, we employ ResNet-18 as the basic encoder (which is also empirically tested in Section \ref{sec: experiment}).
In ResNet-18, there are eight blocks. The first six blocks require convolution operations to align their channel sizes with the final output dimension \( C_{\text{out}} \), as the last two blocks inherently match this size.
	
In the Feature Transformation Layer, the convolution computation for unifying the channel size of the first six nodes into the final size $C_{\text{out}}$ is $\mathcal{O}\big(\sum_{i=1}^6 C_{\text{out}} H_i W_i (C_i + 1)\big)$~\cite{krizhevsky2012imagenet}. 
Similarly, interpolating nodes with varying widths and heights into a unified size incurs a complexity of $\mathcal{O}\big(9N C_{\text{out}} H_{\text{out}} W_{\text{out}}\big)$, accounting for two rounds of interpolations on the values of each pixel's four nearest neighboring points, including a total of nine operations (six multiplications and three additions).
	
In the GAT process, the computational complexity for the two learnable representation transformations $\mathbf{P}^m \in \mathbb{R}^{C_{\text{out}} \times C_{\text{out}}}$ and $\mathbf{U}^m \in \mathbb{R}^{C_{\text{out}} \times C_{\text{out}}}$ is $\mathcal{O}\big(N C_{\text{out}}^2\big)$ each, since in this process each output element is the sum of the products of every element of the input tensor and the corresponding weight. Therefore, for each output element, \(C_{\text{out}}\) multiplications are required. 
Furthermore, calculating attention between the nodes incurs a computational complexity of $\mathcal{O}\big(N(N-1) C_{\text{out}}\big)$, as each node requires the computation of attention values with all other nodes. Multiplying the obtained attention weights with each node results in $\mathcal{O}\left(N^2 C_{\text{out}}\right)$, as each node needs to update its features using the attention weights obtained from other nodes.

In the Relation Mapping Layer, the computational complexity of generating linear weights is $\mathcal{O}(2N T C_{\text{out}} C_{\text{rm}})$, where 
\( C_{\text{rm}} \) is the channel size for relation mapping. 
Moreover, for each of the \( N \) nodes and \( T \) tasks, two linear layers with a weight size of \( C_{\text{out}} \times C_{\text{rm}} \) are employed, each requiring \( C_{\text{rm}} \) multiplications per output element. The complexity of linearly weighting a node is $\mathcal{O}(N T C_{\text{out}} H_{\text{out}} W_{\text{out}})$, and the complexity of summing the weights of all tasks and nodes is $\mathcal{O}((N-1) T C_{\text{out}} H_{\text{out}} W_{\text{out}})$, since these operations involve aggregating \( N \) weighted block features across  \( H_{\text{out}} \) and \( W_{\text{out}} \) spatial dimensions.
As a result, the overall computational complexity of the Relation Mapping Layer is  
\(
\mathcal{O}\big(N T C_{\text{out}} (C_{\text{rm}}
+ 
H_{\text{out}} W_{\text{out}})\big)
,
\)
which scales linearly with the number of tasks \( T \).



Given that $C_{\text{out}}$ significantly outweighs 
$N$, $T$, $H_{\text{out}}$, and $W_{\text{out}}$, the overall computational complexity of \textbf{Algorithm~\ref{algorithm:1}} is 
$\mathcal{O}(NC_{\text{out}}^2)$.
The complexity is relatively low, scales linearly with the number of encoding blocks $N$ at the encoder, and is little affected by the number of tasks $T$.

	\begin{algorithm}
		\caption{Graph Attention Inter-block (GAI) Multi-task Semantic Communication Training}
		\begin{algorithmic} 
			
			\REQUIRE $\{\mathbf{F}_i\}$ $(i = 1, \ldots, N)$, the set of feature outputs by $N$ selected blocks of the Encoder. 
			\ENSURE Task-specific transmission feature $\mathbf{z}_t$ for each task $t$. 
			\STATE \textbf{Initial Setup}
			\STATE Set the training to continue until convergence, with a possible early stop based on validation loss to prevent overfitting. 
			\STATE \textbf{Operation of the Feature Transformation Layer}
			\FOR{each block $i$ in Encoder blocks}
			\STATE \textit{\% Channel Dimension Transformation}
			\STATE Apply ConvLayer to unify the channel dimension: $[C_i, W_i, H_i] \rightarrow [C_{\text{out}}, W_i, H_i]$, obtaining $\mathbf{F}_i$; see~\eqref{eq:convlayer},~\eqref{eq:convlayer2}.
			\STATE \textit{\% Pooling Operation}
			\STATE Apply Pooling to simplify the feature dimension: $[C_{\text{out}}, W_i, H_i] \rightarrow [C_{\text{out}}]$, obtaining the initial node representation $\mathbf{V}_i^{0}$; see~\eqref{eq:pooling}.
			\STATE \textit{\% Interpolation}
			\STATE Apply InterpolateLayer for resizing: $[C_{\text{out}}, W_i, H_i] \rightarrow [C_{\text{out}}, W_{\text{out}}, H_{\text{out}}]$, obtaining $\mathbf{K}_i$; see~\eqref{eq:interpolation}.
			\ENDFOR
			\medskip
			\STATE \textbf{Operation of the Graph Attention Layer}
			\FOR{each graph iteration $m$ $(m = 1$ to $M)$}
			\FOR{each node pair $(i, j)$}
			\STATE Compute attention coefficients $a_{i,j}^m$ using $\mathbf{U}^m$, $\mathbf{V}_i^{m-1}$, and $\mathbf{V}_j^{m-1}$; see~\eqref{eq:attention}.
			\ENDFOR
			\STATE Update the representation of node $i$, i.e., $\mathbf{V}_i^{m}$, based on neighboring nodes and attention coefficients; see~\eqref{eq:updating}.
			\ENDFOR
			\STATE Derive the final node representations \({\mathbf{V}_{\text{1}}^M, \mathbf{V}_{\text{2}}^M, \cdots, \mathbf{V}_{\text{N}}^M}\).

			\medskip
			
			\STATE \textbf{Operation of the Relation Mapping Layer}
			\FOR{each task $t$ $(t = 1$ to $T)$}
			\FOR{each node $i$ $(i = 1$ to $N)$}
			\STATE Process node $i$ through relation mapping layer to generate task-node weights $\mathbf{e}_{i,t}$; see~\eqref{eq:nodeencoding}.
			\ENDFOR
			\STATE Combine the normalized weights $\mathbf{e}_{i,t}$ with feature $\mathbf{K}_i$ to derive the transmission feature $\mathbf{z}_t$ for task $t$; see \eqref{eq:finaltaskencoding}.
			\ENDFOR
			
		\end{algorithmic}
		\label{algorithm:1}
	\end{algorithm}

	\section{Experimental Results}\label{sec: experiment}
	In this section, we conduct experiments on various datasets to assess the efficacy of our proposed GAI module. For an input data sample (e.g., an image and text), the encoder generates features for multiple simultaneous tasks on the data sample (e.g., on the image), such as semantic segmentation, surface normal estimation, depth prediction, keypoint detection, and edge detection; and then transmits the features to the decoders for accomplishing the tasks simultaneously.
	
	\subsection{Experimental Setup}
	\subsubsection{Datasets}
	We validate our experiments using five widely recognized public datasets: CityScapes 2Task \cite{the}, NYU v2 \cite{silberman2012indoor},  TaskonomyTiny 5Tasks \cite{taskonomy},  Oxford-IIIT Pet dataset~\cite{parkhi2012cats}, and the MVSA dataset\cite{niu2016mvsa}. 
	
	\par
	\textbf{CityScapes 2Task~\cite{the}:}
	The CityScapes 2Task dataset offers a diverse collection of urban scenes specifically curated for semantic segmentation and depth prediction tasks. Regarding semantic segmentation, the dataset employs two key evaluation metrics: mean Intersection over Union (mIoU) and pixel accuracy (Pixel Acc), both desirable when higher. mIoU measures the mean ratio of the intersection to the union of the predicted and actual areas of a category. Pixel accuracy is the ratio of correctly classified pixels to the total number of pixels in the image.
	For the depth prediction tasks, the evaluation involves absolute error (Abs), relative error (Rel), and threshold accuracy (\( \delta \)). Abs represents the direct difference between the predicted and actual values, while Rel is the proportion of this difference to the actual value. Another metric in depth prediction is quantified using \( \delta \), which measures the relative difference between the predicted depth value \( d \) and the actual depth value \( d_{gt} \). Specifically, \( \delta \) is defined as \( \delta = \max\left(\frac{d}{d_{gt}}, \frac{d_{gt}}{d}\right) \), and this metric is determined by the proportion of depth estimates where \( \delta \) falls below a certain threshold (\text{thr}). These thresholds are set to \(1.25\), \(1.25^2\), and \(1.25^3\) in the dataset.
	
	\textbf{NYU v2 \cite{silberman2012indoor}:} 
	This dataset is designed to handle three tasks: semantic segmentation, surface normal estimation, and depth prediction. The evaluation metrics for semantic segmentation and depth prediction are consistent with those used on the CityScapes 2Task. For the evaluation of surface normal estimation, the metrics include the mean (Mean) and median (Median) angular distances between the predicted and actual values. Additionally, we calculate the proportion of pixels where the angular difference between the predicted and the ground truth is less than the thresholds of 11.25°, 22.5°, and 30°. Experiments are conducted in NYU v2 2Task (semantic segmentation and surface normal estimation) and NYU v2 3Task (semantic segmentation, surface normal estimation, and depth prediction).
	
	\textbf{TaskonomyTiny 5Task \cite{taskonomy}:}
	This dataset encompasses annotations for a total of 26 tasks. Building upon insights from existing literature, such as \cite{ada} and \cite{rahimian2023dynashare}, we focus on five tasks: semantic segmentation, depth prediction, surface normal estimation, keypoint detection, and edge detection. Our training data comprises samples from 25 scenarios, and the validation and test datasets are drawn from five separate scenes. To evaluate model performance, we employ task-specific loss metrics, as detailed in Section~\ref{Objective Function}.
	
	\textbf{Oxford-IIIT Pet Dataset\cite{parkhi2012cats}:} 
		This dataset comprises 7,349 images of 37 different breeds of cats and dogs.
Each image is annotated for both semantic segmentation and classification, providing detailed labels for each animal and its associated breed. For the classification task, we use classification accuracy as the metric to evaluate performance.
	
	\textbf{MVSA dataset\cite{niu2016mvsa}:}
		This is a multimodal sentiment analysis dataset containing image-text pairs with manual sentiment annotations collected from Twitter. It is commonly used for sentiment analysis tasks that involve both visual and textual data. We utilize the MVSA-single subset, which consists of individual image-text pairs labeled with sentiment categories, for both training and testing. We focus on three tasks: image sentiment classification, text sentiment classification, and multimodal sentiment classification combining both text and images, with classification accuracy as the evaluation metric.
	
	\subsubsection{Metrics} 
	To comprehensively compare the overall performance and assess relative improvements across various tasks, we calculate the average improvement for each task across all metrics. This is quantified by  
	\begin{equation}
		\Delta_{t} = \frac{1}{|G|} \sum_{j=1}^{|G|} l_j \left( \frac{G_{t,j} - G_{st,j}}{G_{st,j}} \right) \times 100\%,
	\end{equation}
	where $|G|$ is the total number of metrics considered for each task $t$. If a smaller value indicates a better performance for metric $G_j$, $l_j=-1$; otherwise, $l_j=1$. Let $G_{st,j}$ denote the score of the $j$-th metric for the $t$-th task in a single-task model.
	
	Then, we derive the aggregate performance improvement, denoted by $\Delta$, by averaging the improvements across all tasks:
	\begin{equation}
		\Delta = \frac{1}{T} \sum_{t=1}^T \Delta_{t}.
	\end{equation}
	
	\subsubsection{Baselines}
	The considered baseline models include single- and multi-task models. 
	
	\textbf{Single-Task Model:}  
	This model can be viewed as single-task DJSCC\cite{dee}, 
	where the encoder encodes information for only one task. Each task is equipped with its own encoder and decoder; i.e., only a single decoder remains active while the others are disabled in Fig.~\ref{fig_1}. 
	
	\textbf{Basic Multi-Task Framework:}
	Multiple tasks share a common encoder, which extracts features applicable across all tasks, and different decoders are assigned for different tasks, as shown in Fig.~\ref{fig_1}. No GAI module is employed. As a result, the output of the last feature extraction block is transmitted to the receiver via the bandwidth-limited, noisy channel. It represents the prevailing strategy in most current semantic communication systems for multi-task transmission.
	
	\textbf{Feature Sharing Models:} 
	These models leverage mechanisms, such as linear layers, gating mechanisms, convolutional modules, and attention mechanisms, to integrate shared and task-specific features, thereby improving the utilization of encoded features across tasks.
	\begin{itemize} 
		\item Cross-Stitch Network \cite{cro}: In this model, each task has its own encoder. Cross-stitch networks utilize a linear layer to blend features from the separate encoders,  allowing for interconnected feature processing across tasks.
		\item Sluice Networks \cite{lat}: Sluice Networks incorporate gating mechanisms, instead of linear layers (e.g., cross-stitch networks), in the encoder, allowing for more dynamic selection and integration of task-specific features.
		\item NDDR-CNN \cite{nddr}: This model introduces a CNN module for integrating shared and task-specific features.
		\item MTAN \cite{endt}: This model emphasizes capturing task-relevant features using an attention mechanism within a shared encoder, showcasing the effectiveness of focused feature extraction. 
	\end{itemize}
	
	\textbf{Dynamic Network Models:} 
	These models employ adaptive architectures and policy-driven mechanisms to select the optimal encoding structures for different tasks.
	\begin{itemize} 
		\item  DEN \cite{ahn}: The DEN model introduces a supplementary network that dynamically adjusts feature channel strategies in the encoder, using reinforcement learning (RL) principles, to formulate more effective information processing policies for each task.
		\item AdaShare \cite{ada}: Central to this model is the generation of distinct policies for each task. These policies decide on the activation or skipping of various feature-extracting residual blocks, that is, selecting different coding block paths for different tasks. This model is known to offer the best results on NYU v2 3Task and TaskonomyTiny 5Task in the literature.
		\item Dynashare \cite{rahimian2023dynashare}: Following AdaShare, this model considers both task and input transmission instance characteristics for distinct policy generation. This model is known to offer the best performance on both CityScapes 2Task and NYU v2 2Task in the literature.
	\end{itemize}
	
	\subsection{Model Architecture and Training Details}
	The following are the architectural design and training details for our model. 
	
	\subsubsection{Encoder Architecture} 
	For single-modal datasets such as CityScapes 2Task \cite{the}, NYU v2 \cite{silberman2012indoor}, TaskonomyTiny 5Task \cite{taskonomy}, and the Oxford-IIIT Pet dataset \cite{niu2016mvsa}, the encoder comprises a standard ResNet encoder and the proposed GAI module.
		For multi-modal datasets like the MVSA dataset \cite{parkhi2012cats}, the encoder uses ResNet18 to extract image features and BERT to extract text features. These features are then fed into the Transformer model.

	$\bullet$ \textbf{ResNet Encoder:}
	This module utilizes different ResNet architectures tailored to specific datasets. ResNet-18 is employed for the NYU V2 2Task, Oxford-IIIT Pet and  MVSA dataset. ResNet-34 is used for all other datasets.

	$\bullet$ \textbf{GAI Module:}
	The GAI module varies based on the underlying ResNet architecture. It contains eight nodes for ResNet-18 and sixteen nodes for ResNet-34.
	In the Feature Transformation Layer, the convolution process with (kernel size = 1) consistently transforms the input to an output dimension of 512 (the channel number of the last block).
	During node representation updating, both $\mathbf{U}$ and $\mathbf{P}$ have the size of $512 \times 512$, ensuring uniform node dimensions for a GAT process.
	The Relation Mapping Layer comprises two linear network layers. The first linear network layer alters the input channel size to 256 and incorporates a ReLU activation function. The second linear network layer adjusts the output channel size to~512.
	
	\subsubsection{Decoder Architecture}
	Each task has its dedicated decoder tailored to the specific task. These decoders are composed of identical structural units, each comprising three sequential convolutional layers:
	\begin{itemize}
		\item The initial layer is a dilated convolution layer with (kernel size = 3, and channel size = \(512\times 1024\)), with varying dilation rates among different units (6, 12, 18, and 24).
		\item An intermediate convolution layer with (kernel size = 1, and channel size =  \(1024\times 1024\)).
		\item The final convolution layer with (kernel size = 1 and channel size = \(512\times [\text{number of classes for the task}]\)) serves for mapping features to the class space and decoding the task-specific results.
	\end{itemize}

	\subsubsection{Training Details}
	The model is trained under sufficient bandwidth (i.e., $R \geq 1$) and negligible noise (i.e., SNR$ \rightarrow +\infty$), utilizing the Adam optimizer with a batch size 8. The training continues until convergence or early stopping is triggered to achieve optimal performance. The node representation updating iteration number $M$ is set to 1. During testing, the model's performance is assessed at various SNR levels, from $-2$ dB to 14 dB.

	\subsection{Comparison With Baselines}
	Tables~\ref{table:cityscapes-2task} to~\ref{table:mvsa 3task} show the performance of task transmission on different datasets, which include 2-task transmissions on CityScapes, 2-task and 3-task transmissions on NYU v2, 5-task transmissions on TaskonomyTiny, 2-task transmissions on  Oxford-IIIT Pet, and 3-task transmissions on the MVSA. 
	
	\begin{table*}[ht]
		\centering
		\caption{Results on the CityScapes 2-task transmission (semantic segmentation and depth prediction). ``↑'' indicates a higher value is preferred. ``↓'' signifies that a lower value is preferred}
		\footnotesize 
		\setlength{\tabcolsep}{4pt} 
			\begin{tabular}{lcccccccccc}
				\toprule
				Models & \multicolumn{2}{c}{Semantic Seg.} & \multicolumn{5}{c}{Depth Prediction} & \multicolumn{3}{c}{Relative (\%)} \\
				\cmidrule(lr){2-3} \cmidrule(lr){4-8} \cmidrule(lr){9-11}
				& mIoU ↑ & Pixel Acc. ↑ & Abs ↓ & Rel ↓ & $\delta < 1.25$ ↑ & $\delta < 1.25^2$ ↑ & $\delta < 1.25^3$ ↑ &  Seg. & Dep. & Ave. \\
				
				\midrule
				Single-Task & 40.2 & 74.7 & 0.017 & 0.33 & 70.3 & 86.3 & 93.3 & - & - & - \\
				Multi-Task & 37.7 & 73.8 & 0.018 & 0.34 & 72.4 & 88.3 & 94.2 & -3.71 & -0.53 & -2.12 \\ 
				Cross-Stitch\cite{cro} & 40.3 & 74.3 & \textbf{0.015} & \textbf{0.30} & 74.2 & 89.3 & 94.9 & -0.14 & 6.32 & 3.08 \\ 
				Sluice\cite{lat} & 39.8 & 74.2 & 0.016 & 0.31 & 73.0 & 88.8 & 94.6 & -0.83 & 4.01 & 1.59 \\ 
				NDDR-CNN\cite{nddr} & 41.5 & 74.2 & 0.017 & 0.31 & 74.0 & 89.3 & 94.8 & 1.28 & 3.28 & 2.23 \\ 
				MTAN\cite{endt} & 40.8 & 74.3 & \textbf{0.015} & 0.32 & 75.1 & 89.3 & 94.6 & 0.48 & 5.30 & 2.89 \\ 
				DEN\cite{ahn} & 38.0 & 74.2 & 0.017 & 0.37 & 72.3 & 87.1 & 93.4 & -3.07 & -1.65 & -2.36 \\ 
				AdaShare\cite{ada} & 41.5 & 74.9 & 0.016 & 0.33 & 75.5 & 89.9 & 94.9 & 1.75 & 3.81 & 2.78 \\
				Dynashare\cite{rahimian2023dynashare} & 45.0 & 75.3 & 0.016 & 0.33 & 74.8 & 89.5 & 94.7 & 6.37 & 3.50 & 4.93 \\
				SimpAtt & 45.6 & 75.2 & 0.016 & 0.30 & 75.7 & 90.5 & 95.3 & 6.72 & 5.08 & 5.90 \\
				GAI-w (ours) &45.6	&75.3&	\textbf{0.015}&	\textbf{0.30}&	75.5&	90.3&	95.2& 7.1&\textbf{7.1}	&7.1\\
				GAI (ours) &\textbf{46.2}	&\textbf{75.4}&	\textbf{0.015}&	\textbf{0.30}&	\textbf{76.9}&	\textbf{90.7}&	\textbf{95.3}& \textbf{7.92}&	\textbf{7.1}&	\textbf{7.51}\\
				\bottomrule
			\end{tabular}%
		\label{table:cityscapes-2task}
	\end{table*}
	
	\begin{table*}[htbp]
		\centering
		\caption{Results on the NYU v2 2-task transmission (semantic segmentation and surface normal estimation)}
		\footnotesize 
		\setlength{\tabcolsep}{4pt} 
			\begin{tabular}{@{}lcccccccccc@{}}
				\toprule
				Models & \multicolumn{2}{c}{Semantic Seg.} & \multicolumn{5}{c}{Surface Normal Estimation}  & \multicolumn{3}{c}{Relative (\%)} \\
				\cmidrule(lr){2-3} \cmidrule(lr){4-8}  \cmidrule(lr){9-11}
				& mIoU ↑ & Pixel Acc. ↑ & Mean ↓ & Median ↓ & 11.25° ↑ & 22.5° ↑ & 30° ↑ & Seg.&Sn.&Ave. \\
				\midrule
				Single-Task  & 27.8          & 58.5          & 17.3                      & 14.4         & 37.2          & 73.7          & 85.1 & - & - & - \\
				Multi-Task    & 22.6          & 55.0          & 16.9                      & 13.7         & 41.0          & 73.1          & 84.3          & -12.3        & 3.1           & -4.6          \\
				Cross-Stitch\cite{cro}    & 25.3          & 57.4          & 16.6                      & 13.2         & 43.7          & 72.4          & 83.8          & -5.4         & 5.3           & -0.1          \\
				Sluice\cite{lat}         & 26.6          & 59.1          & 16.6                      & 13.0~        & 44.1          & 73.0          & 83.9          & -1.6         & 6             & 2.2           \\
				NDRR-CNN\cite{nddr}      & 28.2          & 60.1          & 16.8                      & 13.5         & 42.8          & 72.1          & 83.7          & 2.1          & 4.1           & 3.1           \\
				MTAN\cite{endt}            & 29.5          & 60.8          & 16.5                      & 13.2         & 44.1          & 72.8          & 83.7          & 5            & 5.7           & 5.4           \\
				DEN\cite{ahn}              & 26.3          & 58.8          & 17                        & 14.3         & 39.5          & 72.2          & 84.7          & -2.4         & 1.2           & -0.6          \\
				AdaShare\cite{ada}         & 29.6          & 61.3          & 16.6                      & 12.9         & 45.0          & 72.1          & 83.2          & 5.6          & 6.2           & 5.9           \\
				DynaShare\cite{rahimian2023dynashare} & \textbf{30.2} & 61.4          & 13.5                      & 10.3         & 53.4          & \textbf{80.9} & \textbf{90.3} & \textbf{6.9} & 21.8          & 14.4          \\
				GAI (ours)        & 29.8          & \textbf{61.7} & \textbf{13.3}             & \textbf{9.7} & \textbf{55.6} & \textbf{80.9} & 89.8          & 6.3          & \textbf{24.1} & \textbf{15.2} \\
				\bottomrule
			\end{tabular}%
		\label{table:nyu v2 2 task}
	\end{table*}
	
	\begin{table*}[htbp]
		\renewcommand{\arraystretch}{1}
		\centering
		\caption{Results on the NYU v2 3-task transmission (semantic segmentation, surface normal estimation and depth prediction)}
		\footnotesize
		\setlength{\tabcolsep}{2pt} 
		\begin{tabular}{@{}lcccccccccccccccc@{}}
			\toprule
			Models & \multicolumn{2}{c}{Semantic Seg.} & \multicolumn{5}{c}{Surface Normal Estimation} & \multicolumn{5}{c}{Depth Prediction} & \multicolumn{4}{c}{Relative (\%)} \\
			\cmidrule(lr){2-3} \cmidrule(lr){4-8} \cmidrule(lr){9-13} \cmidrule(lr){14-17}
			&\thead{mIoU ↑} & \thead{Pixel\\Acc. ↑} & \thead{Mean ↓} & \thead{Median ↓} & \thead{11.25° ↑} & \thead{22.5° ↑} & \thead{30° ↑} & \thead{Abs ↓} & \thead{Rel ↓} & 
			\thead{$\delta$\textless\\1.25 ↑} &
			\thead{$\delta$\textless\\1.25$^2$ ↑} &
			\thead{$\delta$\textless\\1.25$^3$ ↑} &
			\thead{Seg.} & \thead{Sn.} &  \thead{Dep.} &  \thead{Ave.} \\
			\midrule
			Single-Task & 27.5 & 58.9 & 17.5 & 15.2 & 34.9 & 73.3 & 85.7 & 0.62 & 0.25 & 57.9 & 85.8 & 95.7 & - & - & - & - \\
			Multi-Task & 24.1 & 57.2 & 16.6 & 13.4 & 42.5 & 73.2 & 84.6 & 0.58 & 0.23 & 62.4 & 88.2 & 96.5 & -7.62 & 7.47 & 5.17 & 1.68 \\
			Cross-Stitch\cite{cro} & 25.4 & 57.6 & 17.2 & 14.0 & 41.4 & 70.5 & 82.9 & 0.58 & 0.23 & 61.4 & 88.4 & 95.5 & -4.92 & 4.22 & 4.67 & 1.32 \\
			Sluice\cite{lat} & 23.8 & 56.9 & 17.2 & 14.4 & 38.9 & 71.8 & 83.9 & 0.58 & 0.24 & 61.9 & 88.1 & 96.3 & -8.42 & 2.86 & 4.13 & -0.48 \\
			NDDR-CNN\cite{nddr} & 21.6 & 53.9 & 17.1 & 14.5 & 37.4 & 73.7 & 85.6 & 0.66 & 0.26 & 55.7 & 83.7 & 94.8 & -14.98 & 2.90 & -3.53 & -5.21 \\
			MTAN\cite{endt} & 26.0 & 57.2 & 16.6 & 13.0 & 43.7 & 73.3 & 84.4 & 0.57 & 0.25 & 62.7 & 87.7 & 95.9 & -4.17 & 8.67 & 3.76 & 2.75 \\
			DEN\cite{ahn} & 23.9 & 54.9 & 17.1 & 14.8 & 36.0 & 73.4 & \textbf{85.9} & 0.97 & 0.31 & 22.8 & 62.4 & 88.2 & -9.94 & 1.69 & -35.24 & -14.50 \\
			AdaShare\cite{ada} & \textbf{30.2} & \textbf{62.4} & 16.6 & \textbf{12.9} & \textbf{45.0} & 71.7 & 84.0 & 0.55 & \textbf{0.20} & 64.5 & 90.5 & 97.8 & \textbf{7.88} & 9.01 & 10.07 & 8.98 \\
			SimpAtt & 29.0 & 60.5 & 16.6 & 13.0 & 44.5 & 71.9 & 83.2 & 0.56 & 0.21 & 63.8 & 89.7 & 97.2 & 5.45 & 7.12 & 8.11 & 6.89 \\
			GAI-w (ours) & 29.7	& 61.1 &17.0	&13.7 &42.4	&70.7	&82.8	&0.53	&0.21 & 66.4 &90.7 &97.5	& 5.86  &7.01	&	10.5&7.79	\\
			GAI (ours) & \textbf{30.2}	&61.4&	\textbf{16.2}&	13.1&	43.7&	\textbf{73.8}&	85.3&	\textbf{0.51}&	\textbf{0.20}	&\textbf{67.7}&	\textbf{91.6}&	\textbf{97.9}&	6.33&	\textbf{9.33}&	\textbf{12.75}&	\textbf{9.47}\\
			\bottomrule
		\end{tabular}%
	\label{table:nyu v2 3 task}
\end{table*}

\begin{table*}[ht]
	\centering
	\caption{Results on the TaskonomyTiny 5-task transmission (semantic segmentation, surface normal estimation, depth prediction, keypoint and edge detection)}
	\footnotesize 
	\setlength{\tabcolsep}{4pt} 
		\begin{tabular}{lccccccccccccc}
			\toprule
			Models & \multicolumn{5}{c}{} & \multicolumn{6}{c}{Relative (\%)} & \\
			\cmidrule(lr){2-6} \cmidrule(lr){7-12}
			& Seg. ↓  & Sn. ↑ & Dep. ↓  & Key. ↓  & Edg. ↓  & Seg. &  Sn. &  Dep. & Key. & Edg. & Ave. \\
			\midrule
			Single-Task & 0.575 & 0.707 & 0.022 & 0.197 & 0.212 & - & - & - & -& -& - \\
			Multi-Task & 0.596 & 0.696 & 0.023 & 0.197 & 0.203 &  -3.65 & -1.56 & -4.55 & 0.00 & 4.25 & -1.10 \\
			Cross-Stitch\cite{cro} & 0.57 & 0.679 & 0.022 & 0.199 & 0.217 & 0.87 & -3.96 & 0.00 & -1.02 & -2.36 & -1.93 \\
			Sluice\cite{lat} & 0.596 & 0.695 & 0.024 & 0.196 & 0.207 & -3.65 & -1.70 & -9.09 & 0.51 & 2.36 & -2.31 \\
			NDDR-CNN\cite{nddr} & 0.599 & 0.700 & 0.023 & 0.196 & 0.203 & -4.17 & -0.99 & -4.55 & 0.51 & 4.25 & -0.99 \\
			MTAN\cite{endt} & 0.621 & 0.687 & 0.023 & 0.197 & 0.206 & -8.00 & -2.83 & -4.55 & 0.00 & 2.83 & -2.51 \\
			DEN\cite{ahn} & 0.737 & 0.686 & 0.027 & 0.192 & 0.203 & -28.17 & -2.97 & -22.73 & 2.38 & 4.25 & -9.42 \\
			AdaShare\cite{ada} & 0.562 & 0.702 & 0.023 & 0.191 & 0.200 & 2.26 & -0.71 & -4.55 & 3.05 & 5.66 & 1.14 \\
			SimpAtt & 0.467 &0.828&0.022&0.193&0.205&18.7&17.1&0&2.03&3.3&7.56\\
			GAI-w (ours) & 0.516 & 0.825 & 0.022 & 0.193 &0.204  & 10.2 & 16.7 & 0 &2.03  & 3.7 & 6.54 \\
			GAI (ours) & \textbf{0.462}&	\textbf{0.831} &	\textbf{0.021}&	\textbf{0.190} &	\textbf{0.196}&\textbf{19.7}&\textbf{17.5} &\textbf{4.5} & \textbf{3.6}& \textbf{7.5}&\textbf{10.54} \\
			\bottomrule
		\end{tabular}%
	\label{table:taskonomy 5task}
\end{table*}

\begin{table}[ht]

	\caption{Results on the Oxford-IIIT Pet 2-task transmission (semantic segmentation and classification)}
	\centering
	\footnotesize
	\begin{tabular}{lcccccccc}
		\toprule
		\multirow{2}{*}{Models} & \multicolumn{2}{c}{Semantic Seg.} & \multicolumn{1}{c}{Classif.} & \multicolumn{3}{c}{Relative (\%)} \\ \cmidrule(lr){2-3} \cmidrule(lr){4-4} \cmidrule(lr){5-7}
		& mIoU ↑ & Pixel Acc. ↑ & Acc. ↑ & Seg.  & Cls.  & Ave.  \\ \midrule
		Single Model            & \textbf{64.9}  & 86.2       & 68.2 & - & - & - \\               
		Multi-Task              & 59.3  & 83.6       & 64.8 & -5.8 & -5.2 &-5.3 \\ 
		Dynashare\cite{rahimian2023dynashare}                    & 62.8  & 84.8       & 70.7& -2.4 & 3.7 & 0.6 \\ 
		GAI (ours)   & 63.3  & \textbf{85.0}       & \textbf{71.0} & -1.8 & \textbf{4.1} & \textbf{1.2} \\ 
		\bottomrule
	\end{tabular}
\label{tab:pet results}
\end{table}

\begin{table*}[h!]
\centering
\footnotesize
\caption{Results on the MVSA 3-task transmission (text and image and multimodal sentiment classification). Acc (T), MacroAvg (T), and WeightedAvg (T) refer to accuracy, macro average, and weighted average for text sentiment classification. Similarly, Acc (I), MacroAvg (I), and WeightedAvg (I) are for image sentiment classification, and Acc (M), MacroAvg (M), and WeightedAvg (M) are for multimodal sentiment classification}
\resizebox{\textwidth}{!}{%
	\begin{tabular}{lccccccccccc}
		\toprule
		\textbf{Model} & \textbf{Metric} & \textbf{Acc (T) ↑} & \textbf{MacroAvg (T) ↑} & \textbf{WeightedAvg (T) ↑} & \textbf{Acc (I) ↑} & \textbf{MacroAvg (I) ↑} & \textbf{WeightedAvg (I) ↑} & \textbf{Acc (M) ↑} & \textbf{MacroAvg (M) ↑} & \textbf{WeightedAvg (M) ↑} & \textbf{RelativeAcc} \\ 
		\midrule
		\multirow{3}{*}{Multi-Task} & Precision & 0.8 & 0.54 & 0.76 & 0.75 & 0.82 & 0.76 & 0.8 & 0.71 & 0.80 & - \\ 
		& Recall & - & 0.57 & 0.80 & - & 0.62 & 0.75 & - & 0.67 & 0.80 & - \\ 
		& F1-Score & - & 0.56 & 0.78 & - & 0.67 & 0.74 & - & 0.68 & 0.80 & - \\ 
		\midrule
		\multirow{3}{*}{GAI (ours)} & Precision & 0.8 & 0.77 & 0.80 & \textbf{0.82} & 0.73 & 0.82 & \textbf{0.85} & 0.89 & 0.86 & 5.19 \\ 
		& Recall & - & 0.75 & 0.80 & - & 0.67 & 0.82 & - & 0.81 & 0.85 & - \\ 
		& F1-Score & - & 0.75 & 0.80 & - & 0.69 & 0.82 & - & 0.84 & 0.85 & - \\ 
		\bottomrule
	\end{tabular}%
}
\label{table:mvsa 3task}
\end{table*}


Table~\ref{table:cityscapes-2task} shows the experimental results on the two tasks of semantic segmentation and depth prediction on the CityScapes 2Task dataset. GAI exceeds all baseline models with an overall gain of $\Delta=+7.51\%$, marking a 2.71\% improvement compared to the leading model DynaShare. In semantic segmentation, GAI achieves a 1.4\% improvement over the baselines. For depth estimation, our results achieve a 1.13\% improvement over the baselines. While the Cross-Stitch model using a separate encoder for each task exhibits proficiency in the depth estimation metrics (best in Abs metrics), it performs poorly on semantic segmentation. In contrast, the proposed GAI adaptively generates a feature for each task by analyzing the correlation among each coding block and weights between the coding block and task, ensuring effective feature utilization.

Table~\ref{table:nyu v2 2 task} presents the experimental results on the NYU v2 2Task dataset, where two tasks of semantic segmentation and surface normal estimation are executed.
The proposed GAI model outperforms all baselines by achieving $\Delta=+15.2\%$ in overall performance, with an improvement of 0.87\% over the optimal Dynashare model.
Although the performance of the GAI model drops by 0.41\% compared to the best baseline in semantic segmentation, i.e., Dynashare, the GAI model improves by 2.17\% in the surface normal transmission task. As a result, overall, GAI is still the best.
The basic multi-task DJSCC model shows a decrease in average performance compared to single-task models. 

Table~\ref{table:nyu v2 3 task} shows the experimental results on the NYU v2 3Task datasets, where three tasks, i.e., semantic segmentation, depth prediction, and surface normal estimation, are considered.
It is observed that the proposed GAI model outperforms all baselines with $\Delta=+9.47\%$, better than the best baseline by 0.80\%. GAI achieves the best results in 8 out of 12 metrics. 
Additionally, the experiments reveal that for most baselines, the performance in the semantic segmentation task tends to be worse than the single-task transmission models. By contrast, the proposed GAI model can maintain consistent improvements across multiple tasks.

Table~\ref{table:taskonomy 5task} summarizes the experimental results on the TaskonomyTiny 5Task dataset, where the five tasks of semantic segmentation, depth prediction, surface normal estimation, keypoint, and edge detection are executed.
The proposed GAI model surpasses all baselines, achieving an overall enhancement of $\Delta=+10.54\%$. GAI significantly outperforms the best baseline, i.e., Ada~\cite{ada}, achieving the best results across all evaluated metrics by 9.47\%. 
One reason for Ada~\cite{ada} providing only marginal improvements in the scenarios involving more complex tasks is due to its learning strategy: In scenarios involving more tasks, the complexity of learning effective task-specific strategies increases, limiting the ability of Ada to adapt and optimize for each task.
In contrast, our GAI model achieves the best metrics among the five evaluated tasks by implementing a unique encoding strategy for each task.

Moreover, the proposed GAI model outperforms the baseline models on the Oxford-IIIT Pet dataset, as shown in Table~\ref{tab:pet results}. Specifically, the GAI model achieves an mIoU of 63.3\% and Pixel Accuracy of 85.0\% in semantic segmentation (c.f. an mIoU of 62.8\% and Pixel Accuracy of 84.8\% by the second-best Dynashare), and a classification accuracy of 71.0\% (c.f. 70.7\% by Dynashare).
The GAI model also outperforms the baseline models across all three tasks on the MVSA dataset, as demonstrated in Table~\ref{table:mvsa 3task}.
It improves the prediction accuracy
from 75.0\% to 82.0\% in the image sentiment classification task, and from 80.0\% to 85.0\% in the multimodal sentiment classification task. The average prediction accuracy across all tasks is improved from 78\% to 82.5\%.

The GAI model generally demonstrates marked improvements in multi-task semantic communication, highlighting the importance of capturing inter-feature relationships.  
This differs substantially from traditional approaches focusing only on independent, sequential feature extraction. Moreover, effectively capturing and utilizing the inter-feature correlations, alongside establishing relationships between the features and tasks, is critical, as this allows the GAI model to meet the demands of different tasks and improve overall task performance.

\subsection{Ablation Study}

Our existing ablation study has focused on the node representation updating mechanism, as the node representation updating in the Graph Attention Layer plays a key role in the proposed GAI model, capturing and processing the representation of the nodes and their correlations.
We experimentally test the GAI model after disabling the node representation updating process, referred to as GAI-w, on the CityScapes 2Task dataset with the two tasks of semantic segmentation and depth prediction, the NYU v2 2Task dataset with the two tasks of semantic segmentation and surface normal estimation, and the Taskonomy 5Task dataset with the five tasks of semantic segmentation, depth prediction, surface normal estimation, and keypoint and edge detection.
Both the Feature Transformation Layer and the Relation Mapping Layer remain enabled, as they are indispensable components within the GAI model.

The experimental results are summarized in Tables~\ref{table:cityscapes-2task},~\ref{table:nyu v2 3 task}, and~\ref{table:taskonomy 5task} (GAI-w). The performance degradation of GAI-w is more pronounced as the tasks increase, with respective decreases of 0.60\%, 2.45\%, and 4.24\%, compared to the full GAI model. In other words, the importance of updating the node representation rises with the increase in tasks and task complexity.
Moreover, even without the node representation updating, the performance of the GAI model is better than the single-task DJSCC model. This suggests that weighting different node representations still confers a performance advantage despite removing the updating mechanism.

We also test a variant of the proposed GAI model, referred to as \textit{SimpAtt}, by replacing the Graph Attention Layer with a simple attention mechanism. This variant helps us to evaluate the contribution of the graph structure to the GAI model. 
While offering improvements over the baseline models, the \textit{SimpAtt} model is consistently outperformed by the GAI model with the Graph Attention Layer. For instance, on the Taskonomy 5Task dataset, the GAI model achieves an average improvement of 10.54\% over the baseline models, compared to the improvement of 7.56\% gained by the \textit{SimpAtt} model.

Tables~\ref{table:city lr} to \ref{table:taskonomy lr} assess the sensitivity of the GAI model to the learning rate, where the learning rate is set to \(10^{-4}\), \(5 \times 10^{-4}\), and \(10^{-3}\).
The results indicate that selecting the appropriate learning rate is crucial for achieving the best possible outcomes. For instance, when the learning rate is \(10^{-4}\), the GAI model performs better across most tasks, suggesting that lower learning rates can be preferable.

\begin{table*}[htbp]
\centering
\caption{Performance comparison under different learning rates on the CityScapes 2Task dataset}
\footnotesize
\label{table:city lr}
\begin{tabular}{cccccccccccc}
\toprule
Learning Rate & \multicolumn{2}{c}{Semantic Segmentation} & \multicolumn{5}{c}{Depth Prediction}  \\
\cmidrule(lr){2-3} \cmidrule(lr){4-8}
 & mIoU $\uparrow$ & Pixel Acc. $\uparrow$ & Abs $\downarrow$ & Rel $\downarrow$ & $\delta < 1.25 \uparrow$ & $\delta < 1.25^2 \uparrow$ & $\delta < 1.25^3 \uparrow$ \\
\midrule
\(10^{-3}\)  & 44.4 & 74.9  & 0.016  & 0.395  & 71.73 & 85.78 & 92.34 \\
\(5 \times 10^{-4}\)  & 45.6 & 75.18 & 0.015 & 0.315 & 75.78 & 90.10 & 95.07 \\
\(10^{-4}\)  & 46.2 & 75.4  & 0.015  & 0.300  & 76.90 & 90.70 & 95.30 \\
\bottomrule
\end{tabular}
\end{table*}

\begin{table*}[htbp]
\centering
\footnotesize  
\caption{Performance comparison under different learning rates on the NYUv2 3Task dataset}
\label{table:nyu lr}
\resizebox{\textwidth}{!}{  
\begin{tabular}{cccccccccccccccc}
	\toprule
Learning Rate & \multicolumn{2}{c}{Semantic Segmentation} & \multicolumn{5}{c}{Surface Normal Estimation} & \multicolumn{5}{c}{Depth Prediction} \\
\cmidrule(lr){2-3} \cmidrule(lr){4-8} \cmidrule(lr){9-13}
 & mIoU $\uparrow$ & Pixel Acc. $\uparrow$ & Mean $\downarrow$ & Median $\downarrow$ & 11.25$^\circ \uparrow$ & 22.5$^\circ \uparrow$ & 30$^\circ \uparrow$ & Abs $\downarrow$ & Rel $\downarrow$ & $\delta < 1.25 \uparrow$ & $\delta < 1.25^2 \uparrow$ & $\delta < 1.25^3 \uparrow$ \\
\midrule
\(10^{-3}\) & 28.6 & 60.5 & 17.1 & 13.3 & 43.8 & 71.1 & 82.3 & 0.56 & 0.21 & 62.8 & 89.3 & 97.3 \\
\(5 \times 10^{-4}\) & 26.5 & 58.8 & 16.8 & 14.2 & 39.8 & 72.4 & 84.9 & 0.56 & 0.22 & 62.7 & 89.1 & 97.2 \\
\(10^{-4}\) & 30.2 & 61.4 & 16.2 & 13.1 & 43.7 & 73.8 & 85.3 & 0.51 & 0.20 & 67.7 & 91.6 & 97.9 \\
\bottomrule
\end{tabular}
}  
\end{table*}

\begin{table}[htbp]
\centering
\footnotesize  
\caption{Performance comparison under different learning rates on the tasknonmy dataset}
\label{table:taskonomy lr}
\begin{tabular}{cccccc}
\toprule
Learning Rate & Depth Abs Err & SN Simi & Keypoint Err & Edge Err & Seg Err \\ 
\midrule
\(10^{-3}\) & 0.0270 & 0.8182 & 0.1926 & 0.2006 & 0.5235 \\ 
\(5 \times 10^{-4}\)& 0.0266 & 0.8196 & 0.1917 & 0.1991 & 0.5020 \\ 
\(10^{-4}\) & 0.0210 & 0.8310 & 0.1900 & 0.1960 & 0.4620 \\ 
\bottomrule
\end{tabular}
\label{tab:hyperparameters}
\end{table}

\subsection{Impact of Bandwidth Ratio}

We investigate the impact of the bandwidth ratio $R$. To support different bandwidth ratios, a convolution module is concatenated at the transmitter’s end to downscale the number of channels from $C_\text{out}$ to  $C_\text{ds}$, and another convolution module is concatenated at the receiver’s end to upscale them back from $C_\text{ds}$ to  $C_\text{out}$.
Here, $C_\text{ds}$ represents the downscaled channel size. For a fair comparison, we 
apply these convolution modules to the proposed GAI model and the baselines when the channel bandwidth is insufficient. The considered models is evaluated under \( R = \frac{1}{4} \), \( \frac{1}{6} \), \( \frac{1}{12} \), and \( \frac{1}{192} \).

It is shown in Tables~\ref{table:cityscapes-r} and~\ref{table:NYU v2 3Task-r} that the proposed GAI model consistently outperforms the baselines in all metrics under all considered \( R \) values.
For the CityScapes 2Task dataset, we observe a marked reduction in the performance of the baseline at \( R=\frac{1}{12} \). While the GAI model also experiences a slight performance decrease compared to the situation with sufficient channel bandwidth, it maintains a significant advantage over the baselines with a performance improvement of \( \Delta=+11.4\% \). On the NYU v2 3Task dataset with three tasks considered, we notice that the relative improvement of the proposed GAI model increases progressively as \( R \) decreases, indicating its effectiveness in scenarios with constrained channel bandwidths.


Tables~\ref{table:cityscapes-r} and~\ref{table:NYU v2 3Task-r} also evaluate the robustness of the proposed GAI model under extremely low bandwidth settings, i.e., 
\( R = \frac{1}{192} \). 
Compared to higher bandwidth ratios, the GAI model exhibits noticeable performance degradation. 
On the CityScapes 2Task dataset, the mIoU of the GAI model for the semantic segmentation task drops from 45.6\% to 41.7\%, and the pixel accuracy decreases from 74.7\% to 74.6\%.
On the depth prediction task, the absolute error increases from 0.015 to 0.020, and the relative error rises from 0.31 to 0.37.
On the NYUv2 3Task dataset, 
the mIoU drops significantly from 29.1\% to 16.7\%, and the pixel accuracy declines from 60.6\% to 52.9\% for semantic segmentation. 
The surface normal estimation task also shows a slight increase in error, and both the Angular Mean and Angular Median increase.
 
Nonetheless, the GAI model still outperforms the best-performing benchmark, i.e., the AdaShare model, in most tasks even under \( R = \frac{1}{192} \). Notably, the mIoU and depth prediction accuracy of the GAI model remain superior to those of the AdaShare model on the CityScapes dataset. 
On the NYUv2 dataset, the GAI model outperforms the AdaShare model in semantic segmentation, surface normal estimation, and depth prediction, e.g., in absolute and relative errors.

This consistent performance under bandwidth constraints demonstrates that the GAI model is capable of efficiently utilizing the limited computing and communication resources by prioritizing the transmission of critical features. This ability enhances the overall task performance and the model's adaptability to varying operational conditions, making it valuable in practical applications where bandwidth is often limited.

\begin{table*}[ht]
\centering
\caption{Comparison of the performance of GAI and the AdaShare model under different bandwidth ratios $R$ in the CityScapes 2Task dataset}
\footnotesize 
\setlength{\tabcolsep}{4pt} 
	\begin{tabular}{lcccccccccc}
		\toprule
		Models & \multicolumn{2}{c}{Semantic Seg.} & \multicolumn{5}{c}{Depth Prediction} & \multicolumn{3}{c}{Relative (\%)} \\
		\cmidrule(lr){2-3} \cmidrule(lr){4-8} \cmidrule(lr){9-11}
		& mIoU ↑ & Pixel Acc. ↑ & Abs ↓ & Rel ↓ & $\delta < 1.25$ ↑ & $\delta < 1.25^2$ ↑ & $\delta < 1.25^3$ ↑ &  Seg. & Dep. & Ave. \\
		\midrule
		GAI ($R = \frac{1}{4}$) & 45.6 & 74.7 & 0.015 & 0.31 & 75.4 & 89.6 & 94.9 & 5.96 & 1.91 & 3.93 \\
		Ada ($R = \frac{1}{4}$) & 40.6 & 75.0 & 0.016 & 0.32 & 75.2 & 89.4 & 94.7 &  &  &  \\ 
		GAI ($R = \frac{1}{6}$)& 45.2 & 75.1 & 0.017 & 0.30 & 73.6 & 89.6 & 94.9 & 5.87 & -0.38 & 2.75 \\
		Ada ($R = \frac{1}{6}$) & 40.5 & 75.0 & 0.016 & 0.32 & 75.0 & 89.3 & 94.7 &  &  & \\ 
		GAI ($R = \frac{1}{12}$) & 45.1 & 75.0 & 0.014 & 0.30 & 76.2 & 90.3 & 95.3 & 9.93 & 12.87 & 11.40 \\
		Ada ($R = \frac{1}{12}$) & 38.1 & 73.9 & 0.020 & 0.38 & 71.0 & 86.6 & 93.4 &  &  &  \\ 
         GAI ($R = \frac{1}{192}$) & 41.7 & 74.6 & 0.020 & 0.370 & 71.6 & 87.0 & 93.4 & 13.67& 2.50 &8.08 \\
        Ada ($R = \frac{1}{192}$) & 37.4 & 64.4 & 0.020 & 0.36 & 64.4 & 84.2 & 92.7 &  &  &  \\
		\bottomrule
	\end{tabular}%
\label{table:cityscapes-r}
\end{table*}

\begin{table*}[htbp]
\renewcommand{\arraystretch}{1}
\centering
\caption{Comparison of the performance of GAI and the AdaShare model under different bandwidth ratios $R$ in the NYU v2 3Task dataset}
\footnotesize
\setlength{\tabcolsep}{2pt} 
\resizebox{\textwidth}{!}{%
	\begin{tabular}{@{}lcccccccccccccccc@{}}
		\toprule
		Models & \multicolumn{2}{c}{Semantic Seg.} & \multicolumn{5}{c}{Surface Normal Estimation} & \multicolumn{5}{c}{Depth Prediction} & \multicolumn{4}{c}{Relative (\%)} \\
		\cmidrule(lr){2-3} \cmidrule(lr){4-8} \cmidrule(lr){9-13} \cmidrule(lr){14-17}
		& \thead{mIoU ↑} & \thead{Pixel\\Acc. ↑} & \thead{Mean ↓} & \thead{Median ↓} & \thead{11.25° ↑} & \thead{22.5° ↑} & \thead{30° ↑} & \thead{Abs ↓} & \thead{Rel ↓} & 
		\thead{$\delta$\textless\\1.25 ↑} &
		\thead{$\delta$\textless\\1.25$^2$ ↑} &
		\thead{$\delta$\textless\\1.25$^3$ ↑} &
		\thead{Seg.} & \thead{Sn.} &  \thead{Dep.} &  \thead{Ave.} \\
		\midrule
		GAI ($R = \frac{1}{4}$)  & 29.1           & 60.6      & 16.4                         & 12.9 & 44.6  & 72.8 & 84.1 & 0.53                        & 0.21  &66.4  &90.6       &97.6       & 1.3                   & 3.3  & 4.4   & 3.0     \\
		Ada ($R = \frac{1}{4}$)  & 28.3           & 60.7      & 16.9                         & 13.6 & 43.3  & 70.3 & 82.5 & 0.59                        & 0.21  & 60.2 & 88.5      & 97.0      &                       &      &       &         \\
		GAI ($R = \frac{1}{6}$)  & 29.5           & 60.4      & 16.3                         & 12.8 & 45.0    & 73.2 & 84.4 & 0.53                      & 0.22  & 66.7 & 90.6      & 97.5      & 1.94                  & 3.16 & 4.42  & 3.17    \\
		Ada ($R = \frac{1}{6}$)  & 28.4           & 60.3      & 16.7                         & 13.5 & 43.2  & 71.3 & 83.4 & 0.59                        & 0.22  & 61.3 & 88.7      & 96.9      &                       &      &       &         \\
		GAI ($R = \frac{1}{12}$)  & 28.1           & 59.9      & 16.4                         & 13.0   & 44.3  & 72.8 & 84.2 & 0.53                      & 0.22  & 65.9 & 90.2      & 97.4      & 0.71                  & 3.59 & 7.6   & 3.97    \\
		Ada ($R = \frac{1}{12}$)  & 27.8           & 59.7      & 17.0                         & 13.7 & 42.5  & 70.5 & 82.7 & 0.63                        & 0.23  & 58.4 & 86.8      & 96.4      &                       &      &       &   \\
        GAI ($R = \frac{1}{192}$) & 16.7 & 52.9 & 16.9 & 13.7 & 41.0 & 72.9 & 84.6 & 0.54 & 0.21 & 65.0 & 90.4 & 97.6 & 3.4 & 1.78 & 6.78 & 3.98 \\
         Ada ($R = \frac{1}{192}$) & 16.3 & 50.7 & 17.2 & 14.1 & 40.4 & 71.4 & 84.0 & 0.62 & 0.22 & 57.9 & 87.4 & 96.9 &  &  &  &  \\
		\bottomrule
	\end{tabular}%
}
\label{table:NYU v2 3Task-r}
\end{table*}

\subsection{Performance Under Different Channel Conditions}

\begin{figure}[t]
\centering
\begin{subfigure}{0.24\textwidth}
	\centering
	\includegraphics[width=4.2cm]{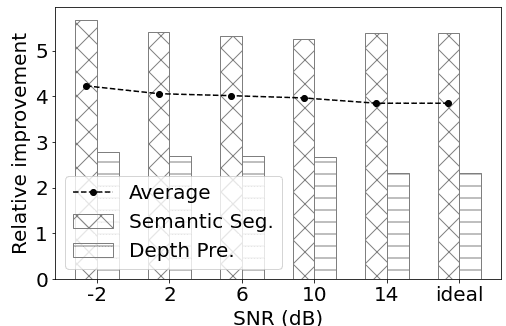}	\caption{CityScapes 2Task}\label{fig:city_snr_1}
\end{subfigure}
\begin{subfigure}{0.24\textwidth}
	\centering
	\includegraphics[width=4.5cm]{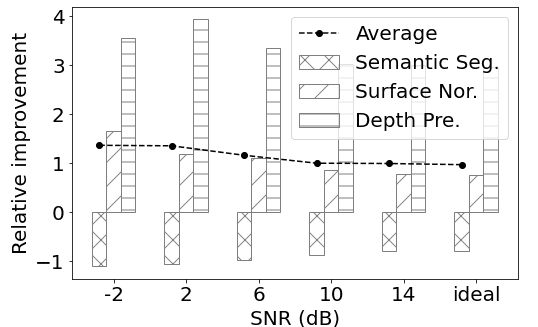}	\caption{NYU v2 3Task}\label{fig:nyu_snr_1}
\end{subfigure}
\caption{Improvement of GAI over Ada under different SNR conditions in the \textbf{CityScapes 2Task} and \textbf{NYU v2 3Task} transmission.}
\end{figure}

To evaluate the performance of the GAI model relative to the baselines under different SNRs, we conduct experiments on the CityScapes 2Task dataset and NYU v2 3Task dataset. As shown in Figs.~\ref{fig:city_snr_1} and~\ref{fig:nyu_snr_1}, a bar chart in the graph represents the relative improvement of different tasks, and the line represents the average relative improvement on multiple tasks.
Overall, the GAI model consistently outperforms the baselines across the spectrum of the SNR ranging from  $-2$ dB to $14$ dB on both datasets.
In addition, the relative improvement is slightly higher as the noise increases.

\begin{figure}[t]
\centering
\begin{subfigure}{0.24\textwidth}
	\centering
	\includegraphics[width=4.2cm]{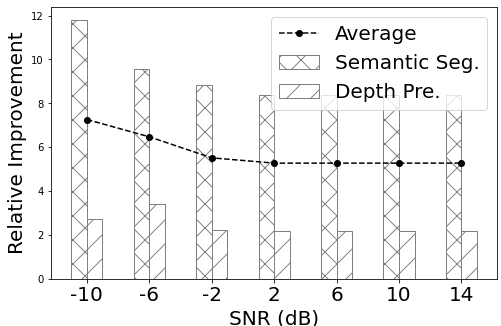}	\caption{CityScapes 2Task}\label{fig:fading_city}
\end{subfigure}
\begin{subfigure}{0.24\textwidth}
	\centering
	\includegraphics[width=4.5cm]{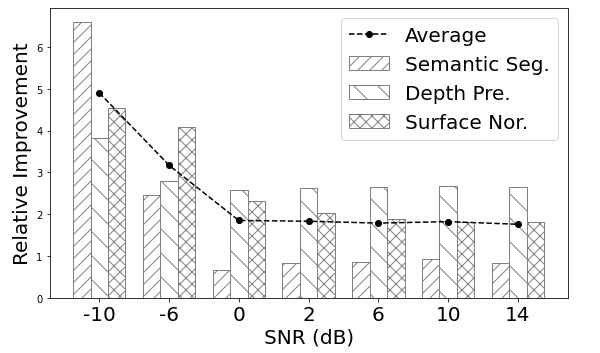}	\caption{NYU v2 3Task}\label{fig:fading_nyu}
\end{subfigure}
\caption{Improvement of GAI over Ada under different SNR conditions in the \textbf{CityScapes 2Task} and \textbf{NYU v2 3Task} transmission over fading channels.}
\end{figure}

We also train and test the GAI model in a Rayleigh fading channel, which simulates the effects of multipath propagation and fast fading commonly encountered in wireless communication environments.
We set the Rayleigh fading parameter to 0.2~\cite{djscc-f}. The GAI model is trained under an SNR of -6 dB and evaluated under the SNRs ranging from -10 dB to 14 dB, using the Cityscapes dataset for 2-task transmissions and the NYU v2 dataset for 3-task transmissions.
It is observed 
that across a wide SNR range from -10 dB to 14 dB, the GAI model significantly outperforms the best-performing benchmark, i.e., AdaShare, on both datasets. On the Cityscapes dataset, the GAI model maintains high performance even in the low SNR regime, particularly in the semantic segmentation task, as shown in Fig.~\ref{fig:fading_city}. On the NYU dataset, the GAI model shows notable performance improvements across all SNR conditions, demonstrating stronger adaptability in tasks, such as depth prediction, as shown in Fig.~\ref{fig:fading_nyu}.

This consistent performance under low SNR conditions confirms the GAI model's ability to leverage feature correlations while adapting task-specific weights effectively. Prioritizing critical features in noisy wireless channels, the model contributes to successfully executing multiple tasks and demonstrates robust performance in non-ideal channel conditions. Such adaptability is crucial in real-world applications, where reliability is paramount, including remote healthcare monitoring and autonomous driving systems.

\subsection{Relationships Among Different Tasks and Encoding Blocks}
\par
\begin{figure*}[t]
\centering
\begin{subfigure}{0.3\textwidth}
	\centering
	\includegraphics[width=5cm]{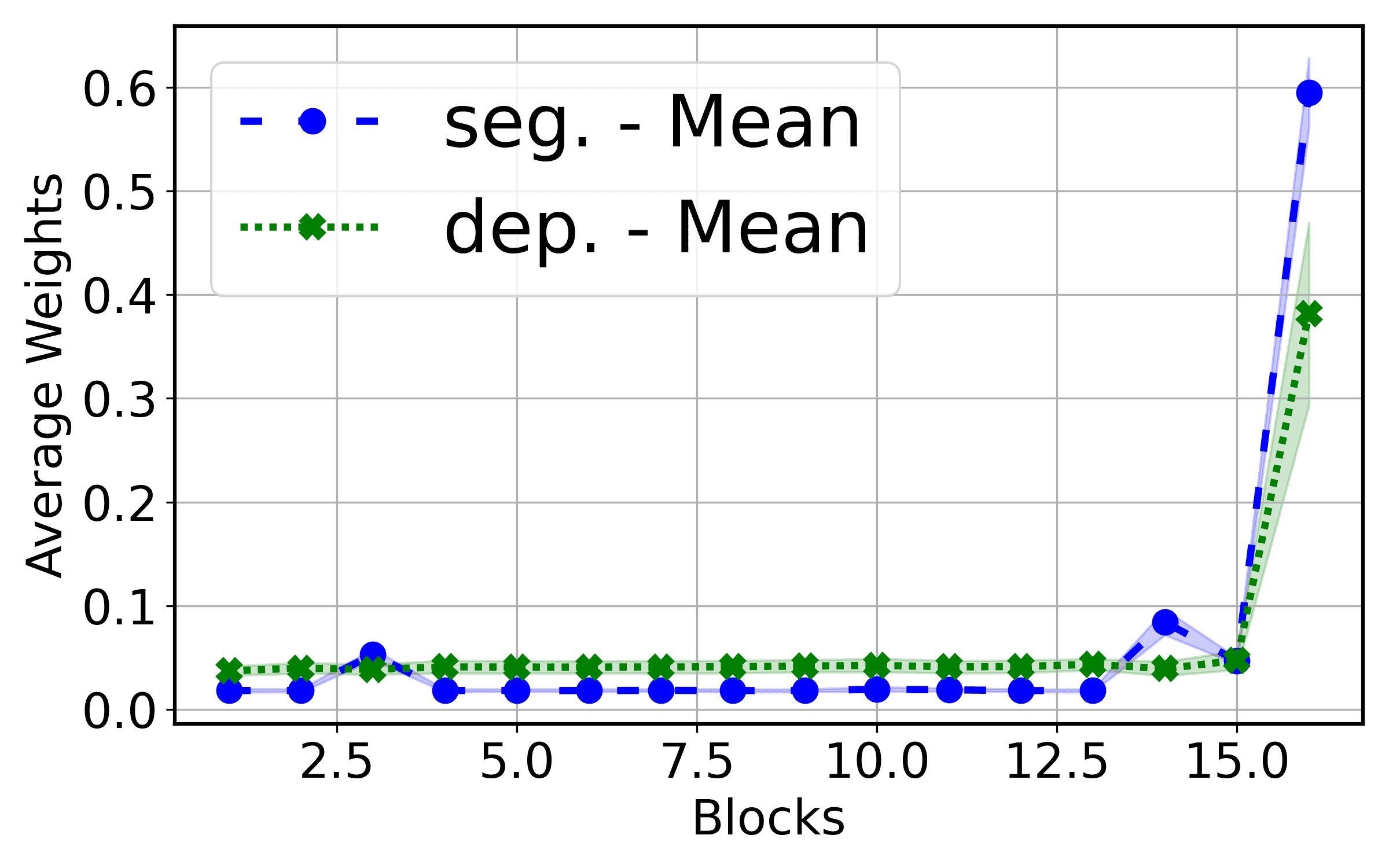}	\caption{CityScapes 2Task}\label{fig:cityscapes-weight}
\end{subfigure}
\begin{subfigure}{0.3\textwidth}
	\centering
	\includegraphics[width=5cm]{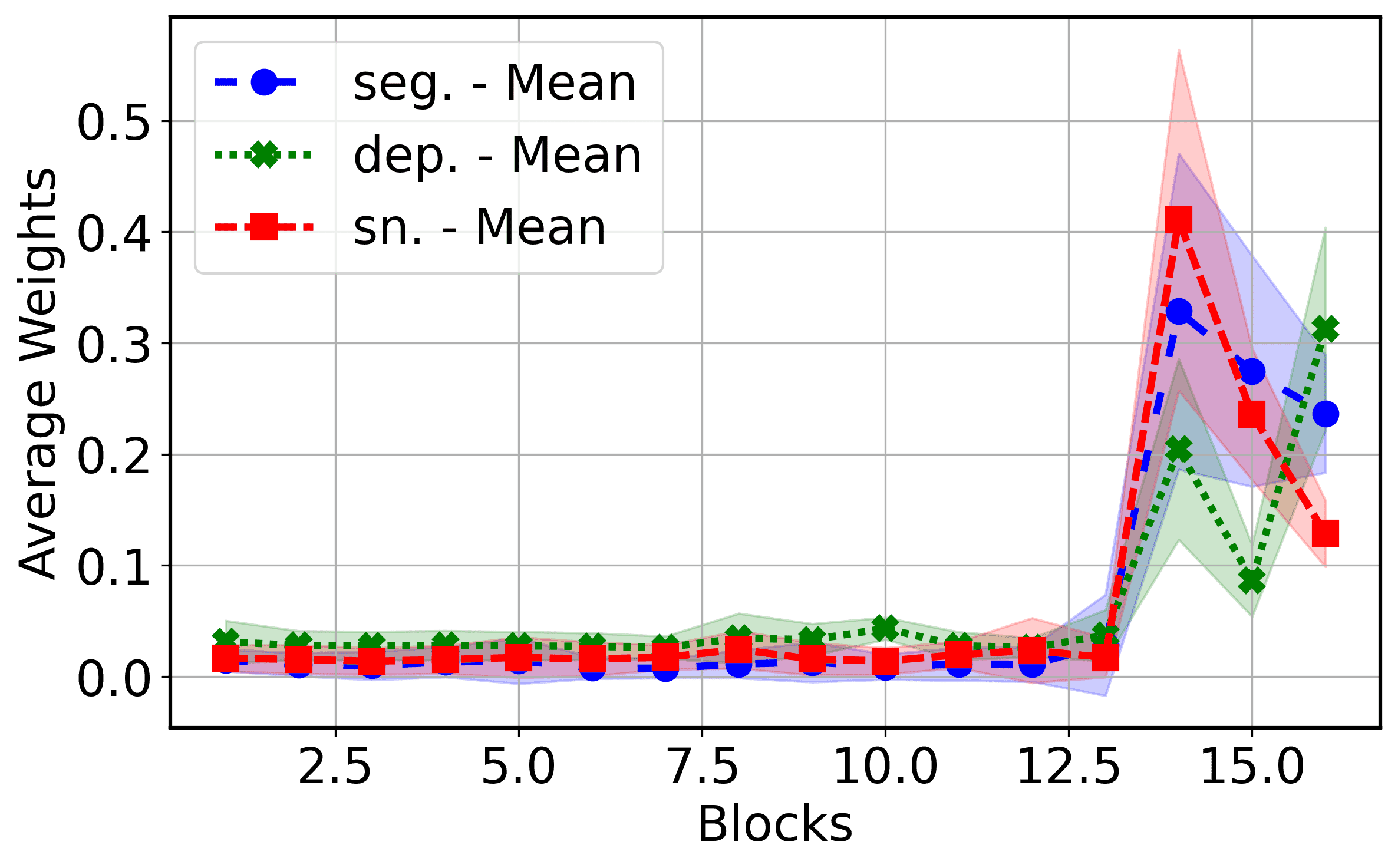}	\caption{NYU v2 3Task }\label{fig:nyuv2-weight}
\end{subfigure}
\begin{subfigure}{0.3\textwidth}
	\centering
	\includegraphics[width=5cm]{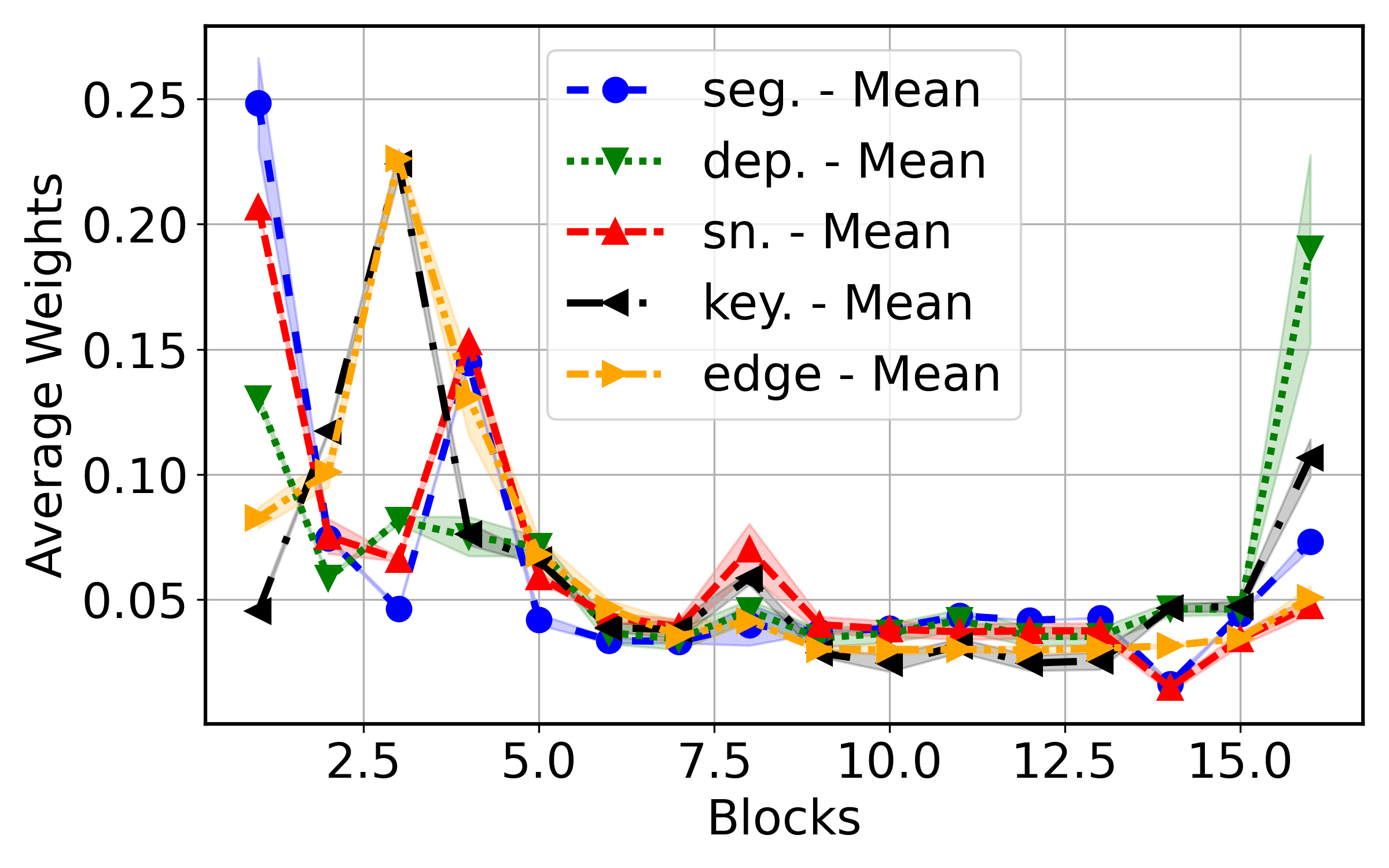}	\caption{TaskonomyTiny 5task }\label{fig:taskonomyTiny-weight}
\end{subfigure}
\caption{The relationship among different tasks and each coding block on the three datasets.}
\end{figure*}

Last but not least, we evaluate the impact of the average weights \({\mathbf{e}_{i,t} \in \mathbb{R}^{C_{\text{out}}}}\) on each task at every node, as demonstrated in Figs.~\ref{fig:cityscapes-weight},~\ref{fig:nyuv2-weight}, and~\ref{fig:taskonomyTiny-weight}. The solid lines represent the average values across different \( R \) values. The shaded areas indicate the variance under different bandwidth ratios. 
Based on the experiments, the following observations are made: 
\begin{itemize}
\item The distinct weight distribution reflects each task requires specific encoding tailored for its unique characteristics. The later blocks in the architecture hold more vital significance across all tasks.
\item 
On the TaskonomyTiny 5Task dataset, specific tasks demonstrate similar demands for the same encoding feature (i.e., form the same feature-extracting blocks). For instance, surface normal estimation and depth prediction, keypoint detection, and edge detection exhibit stronger correlations.

\item As the tasks increase, the complexity of the weight distribution escalates.
On the TaskonomyTiny 5Task dataset, increased reliance on information from shallow layers (closer to the input) is observed across the tasks. In this sense, utilizing information from the shallow layers can be beneficial, as it contains more complete features and is conducive to multi-task execution.
\end{itemize}
\par

\begin{figure}[t]
\centering
\includegraphics[width=0.3\linewidth]{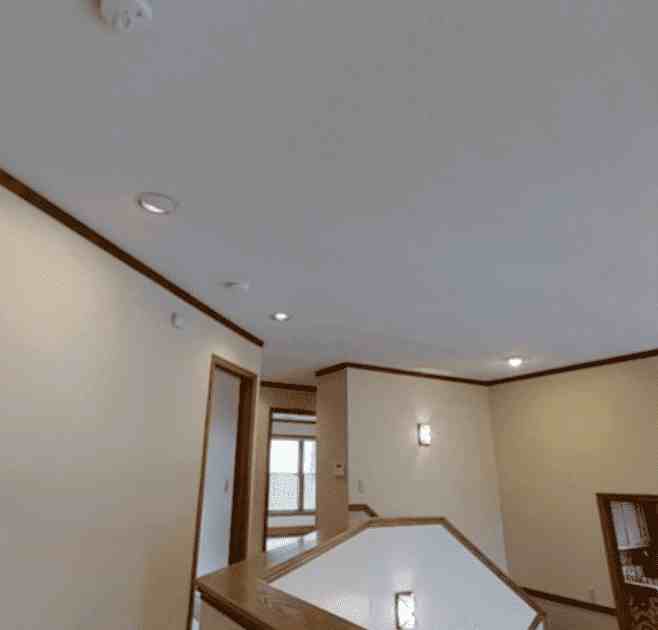}
\caption{The input image.}
\label{fig:input_image}
\end{figure}

\begin{figure*}[!ht]
\centering

\subfloat[block 1]{\includegraphics[width=0.24\linewidth]{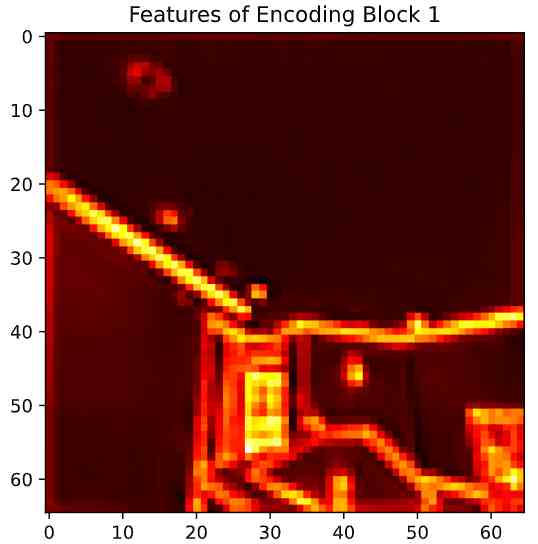}\label{fig:fig_16b1}}%
\hfil
\subfloat[block 2]{\includegraphics[width=0.24\linewidth]{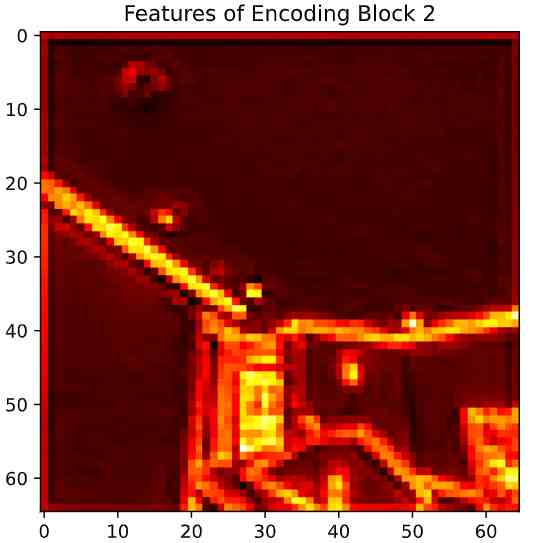}\label{fig_16b2}}%
\hfil
\subfloat[block 3]{\includegraphics[width=0.24\linewidth]{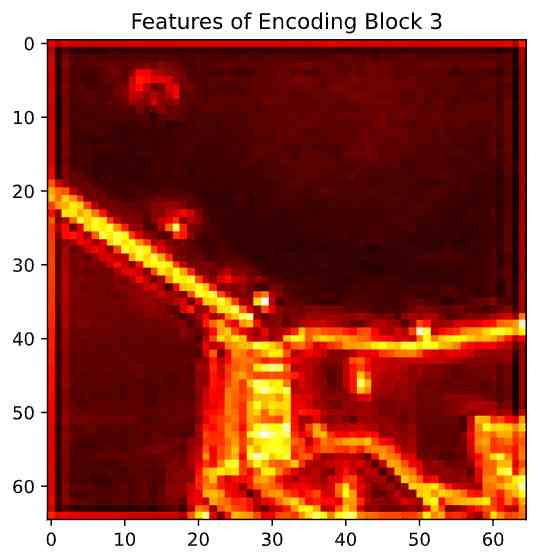}\label{fig_16b3}}%
\hfil
\subfloat[block 4]{\includegraphics[width=0.23\linewidth]{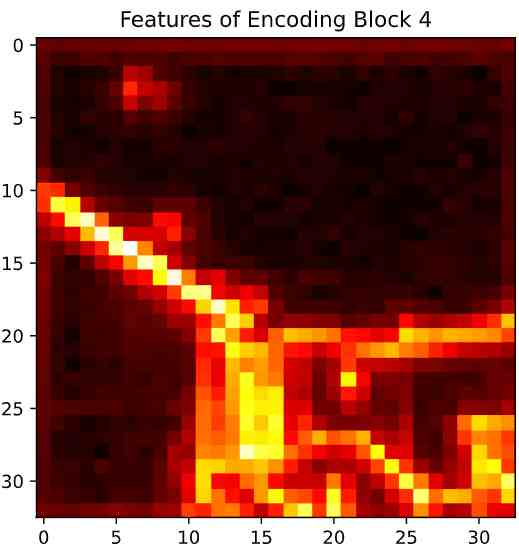}\label{fig_16b4}}%

\vfil 

\subfloat[block 13]{\includegraphics[width=0.24\linewidth]{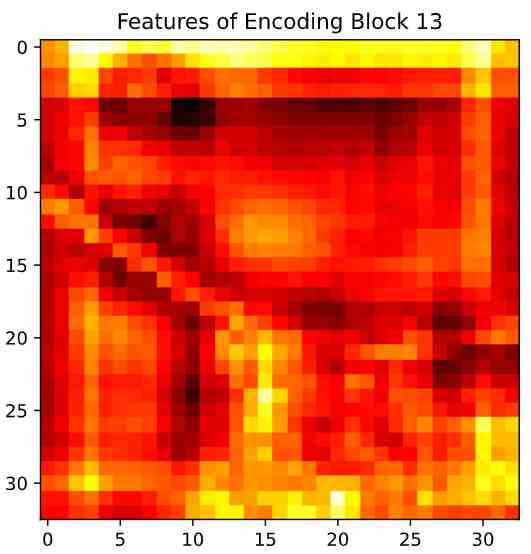}\label{fig_16c1}}%
\hfil
\subfloat[block 14]{\includegraphics[width=0.24\linewidth]{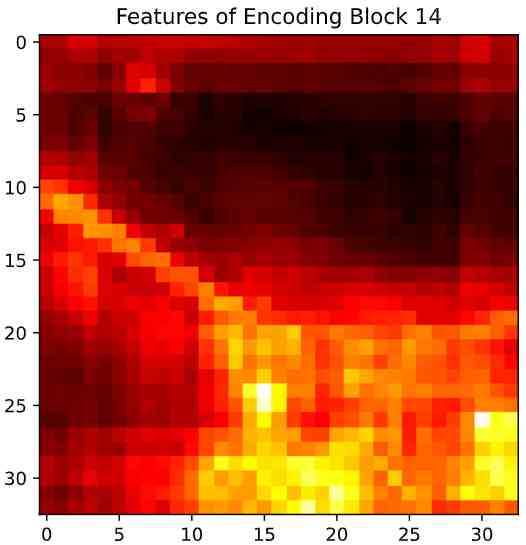}\label{fig_16c2}}%
\hfil
\subfloat[block 15]{\includegraphics[width=0.24\linewidth]{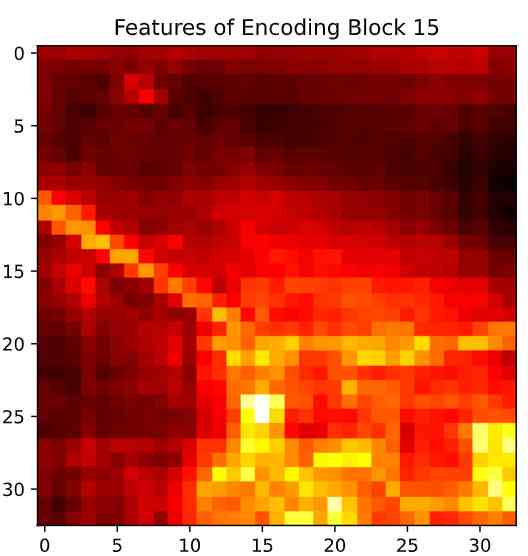}\label{fig_16c3}}%
\hfil
\subfloat[block 16]{\includegraphics[width=0.24\linewidth]{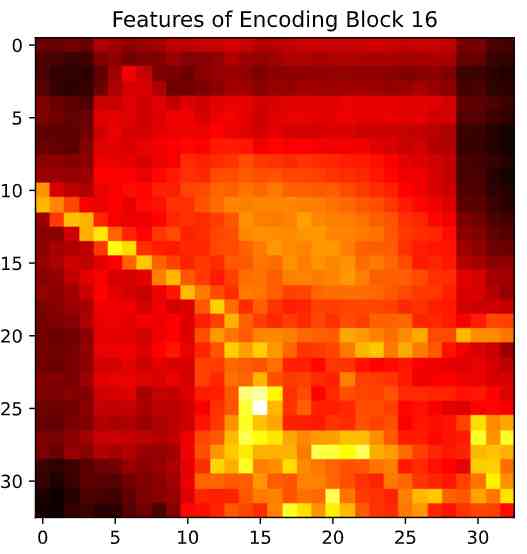}\label{fig_16c4}}%

\vfil 

\subfloat[Feature of task 1]{\includegraphics[width=0.15\linewidth]{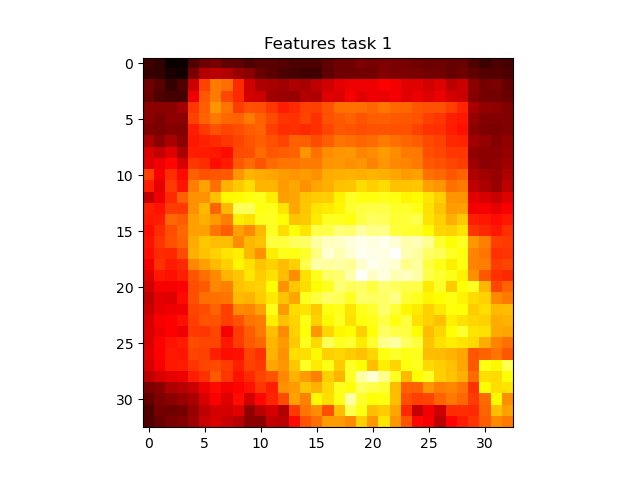}\label{fig_t0}}%
\hfil
\subfloat[Feature of task 2]{\includegraphics[width=0.15\linewidth]{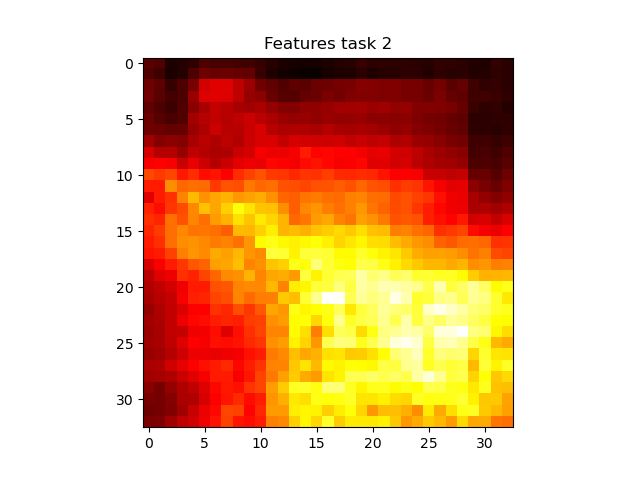}\label{fig_t1}}%
\hfil
\subfloat[Feature of task 3]{\includegraphics[width=0.15\linewidth]{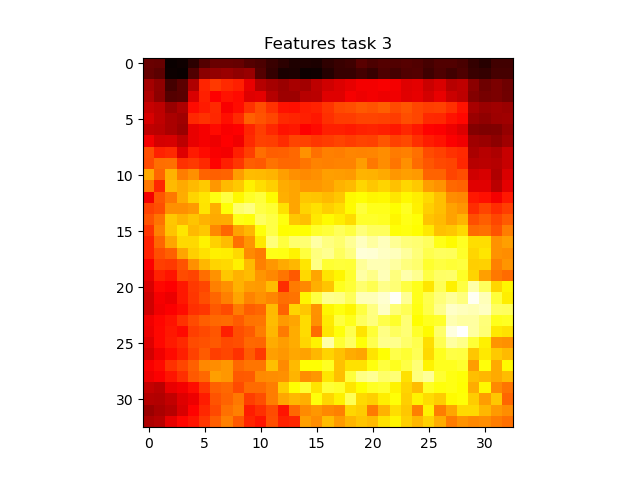}\label{fig_t2}}%
\hfil
\subfloat[Feature of task 4]{\includegraphics[width=0.15\linewidth]{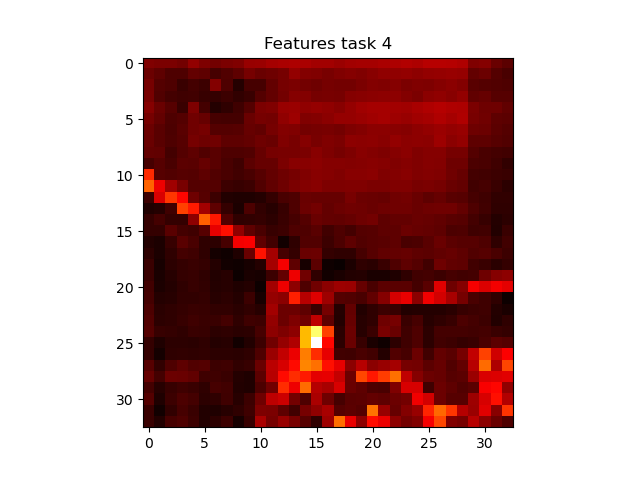}\label{fig_t3}}%
\hfil
\subfloat[Feature of task 5]{\includegraphics[width=0.15\linewidth]{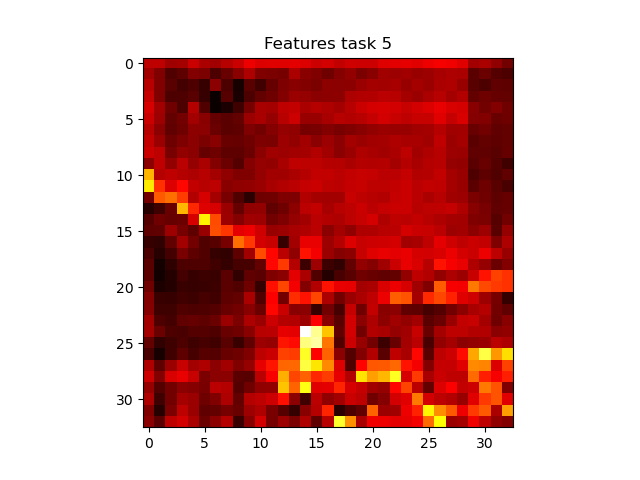}\label{fig_t4}}%

\caption{ (i) Shows the information encoded from the early blocks (1-4), which typically capture basic details such as edges and textures. (ii) Demonstrates the information encoded from the deeper blocks (13-16), where complex, high-level semantic information is encapsulated. (iii) Different transmitted task features.}
\label{fig:overall}
\end{figure*}

A more detailed insight into the feature encoding process from different blocks is shed in Fig.~\ref{fig:overall}. The input image is provided in Fig.~\ref{fig:input_image}. There are a total of 16 encoding blocks. We visualize the encoded features in the first and last four encoding blocks. The blocks closer to the input layer, specifically the first four blocks shown in Figs.~\ref{fig:fig_16b1} to~\ref{fig_16b4}, focus more on local details, such as edges and lines. By contrast, the blocks closer to the output layer, such as the last four blocks depicted in Figs.~\ref{fig_16c1} to~\ref{fig_16c4}, concentrate more on abstract and complex information. As shown in Figs.~\ref{fig_t0} to~\ref{fig_t4}, the transmitted features for different tasks can differ significantly.
The GAI model is capable of tailoring the extracted features to meet the specific demands of various tasks.

\begin{figure*}[htbp]
    \centering
    \includegraphics[width=0.9\textwidth]{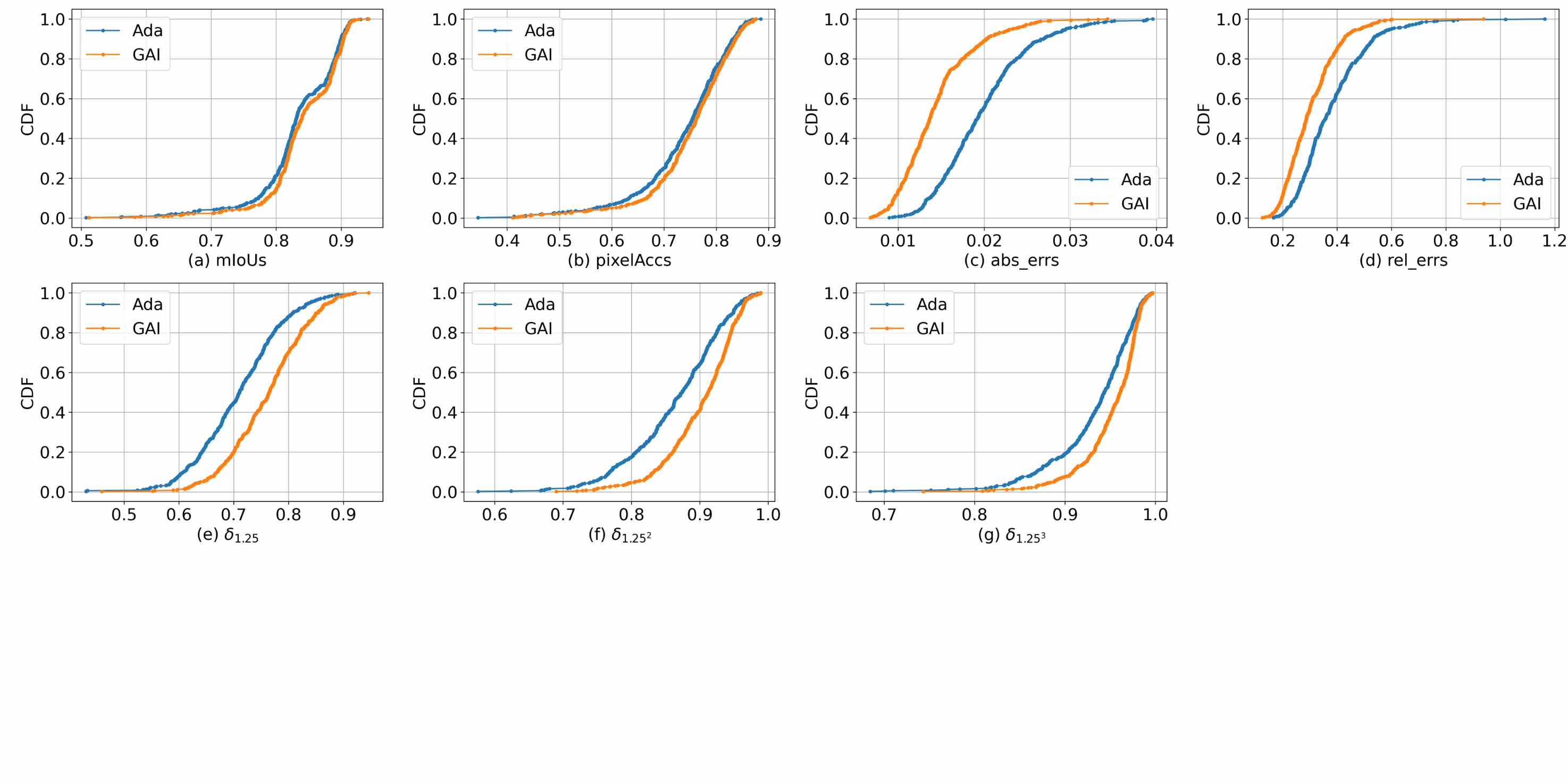}
    \vspace{-20mm}
    \caption{Outage probability in the CityScapes 2Task transmission. For most metrics, curves closer to the bottom-right indicate better performance. Exceptions include ``abs\_errs'' and ``rel\_errs'', where curves closer to the top-left are preferable.}
    \label{fig:city_cdf}
\end{figure*}

\begin{figure*}[htbp]
    \centering
    \includegraphics[width=0.9\textwidth]{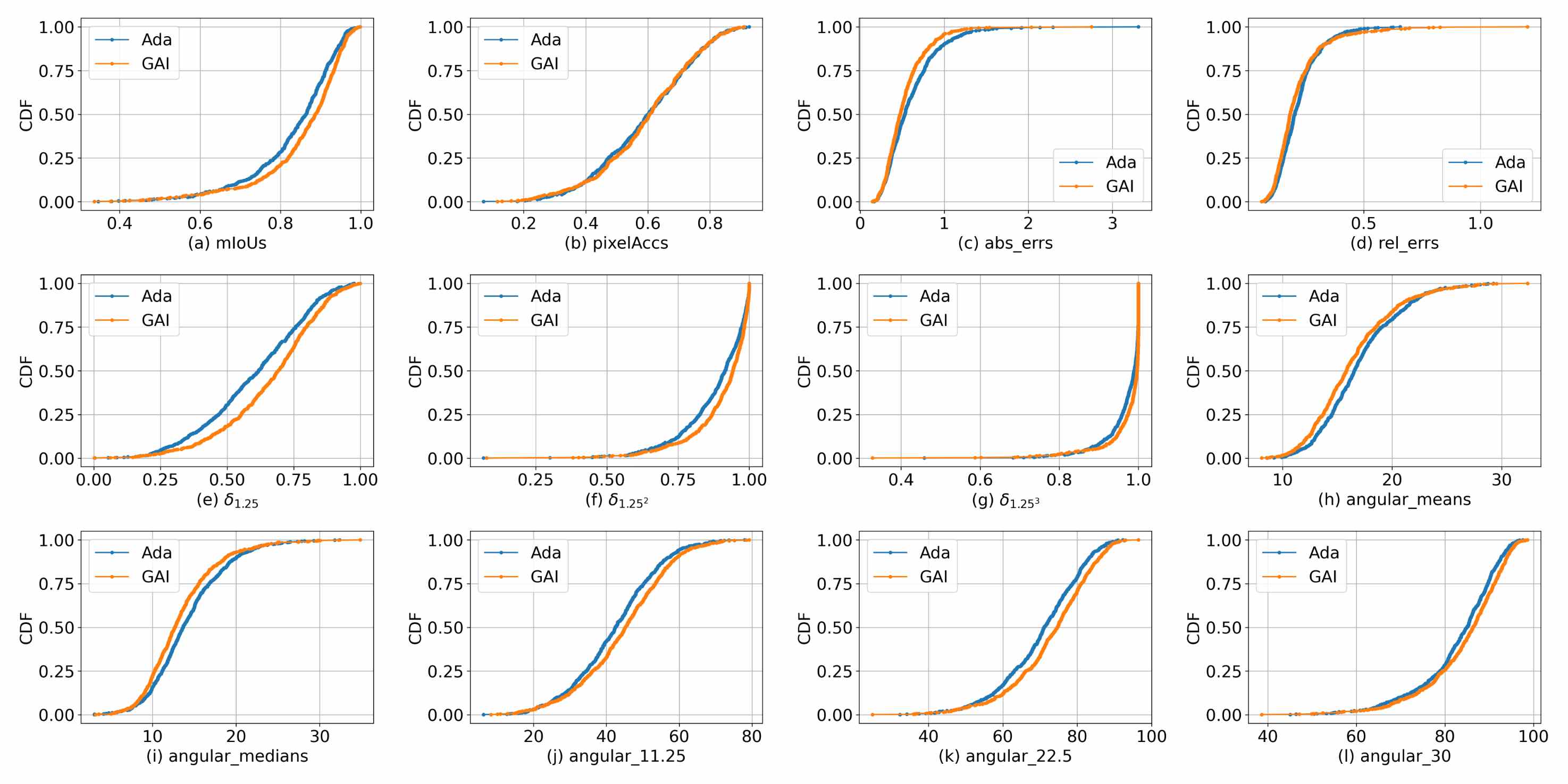}
    \caption{Outage probability in the NYU v2 3-Task transmission. For most metrics, curves closer to the bottom-right indicate better performance. Exceptions include ``abs\_errs'', ``rel\_errs'', ``angular\_means'', and ``angular\_medians'', where curves closer to the top-left are preferable.}
    \label{fig:nyu_cdf}
\end{figure*}

Figs.~\ref{fig:city_cdf} and~\ref{fig:nyu_cdf} compare the cumulative distribution functions (CDFs) of the relevant performance metrics between the proposed GAI model and the best-performing benchmark model, AdaShare, on the Cityscapes and NYU datasets. The GAI model consistently outperforms AdaShare in all evaluated tasks.
    For the \textit{semantic segmentation} task, the CDF curves of the mIoU and pixel accuracy are consistently closer to the bottom-right corner under the GAI model, reflecting higher accuracy and reliability on both datasets.    
    For the \textit{depth prediction} task, the GAI model exhibits superior performance with lower absolute and relative errors.
    For the \textit{surface normal estimation} task on the NYU dataset, the GAI model demonstrates better error control with the CDFs of angle-related metrics closer to the top-left corner, indicating smaller prediction errors.

\begin{figure*}[htbp]
	\centering
	\begin{subfigure}[b]{0.3\textwidth}
		\centering
		\includegraphics[height=3cm, keepaspectratio]{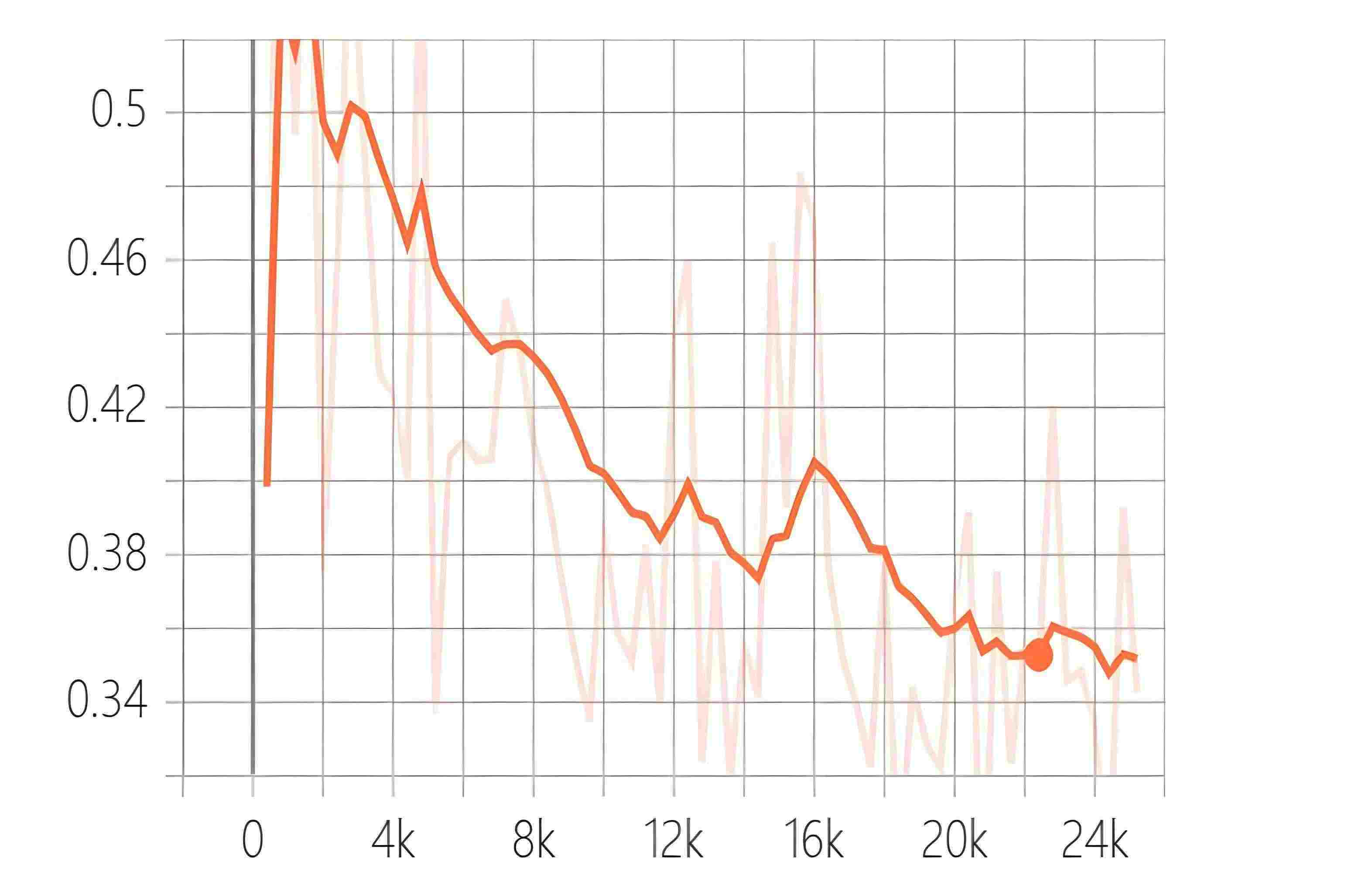}
		\caption{Training trails of Cityscapes}
		\label{fig:city_trails}
	\end{subfigure}
	\hfill
	\begin{subfigure}[b]{0.3\textwidth}
		\centering
		\includegraphics[height=3cm, keepaspectratio]{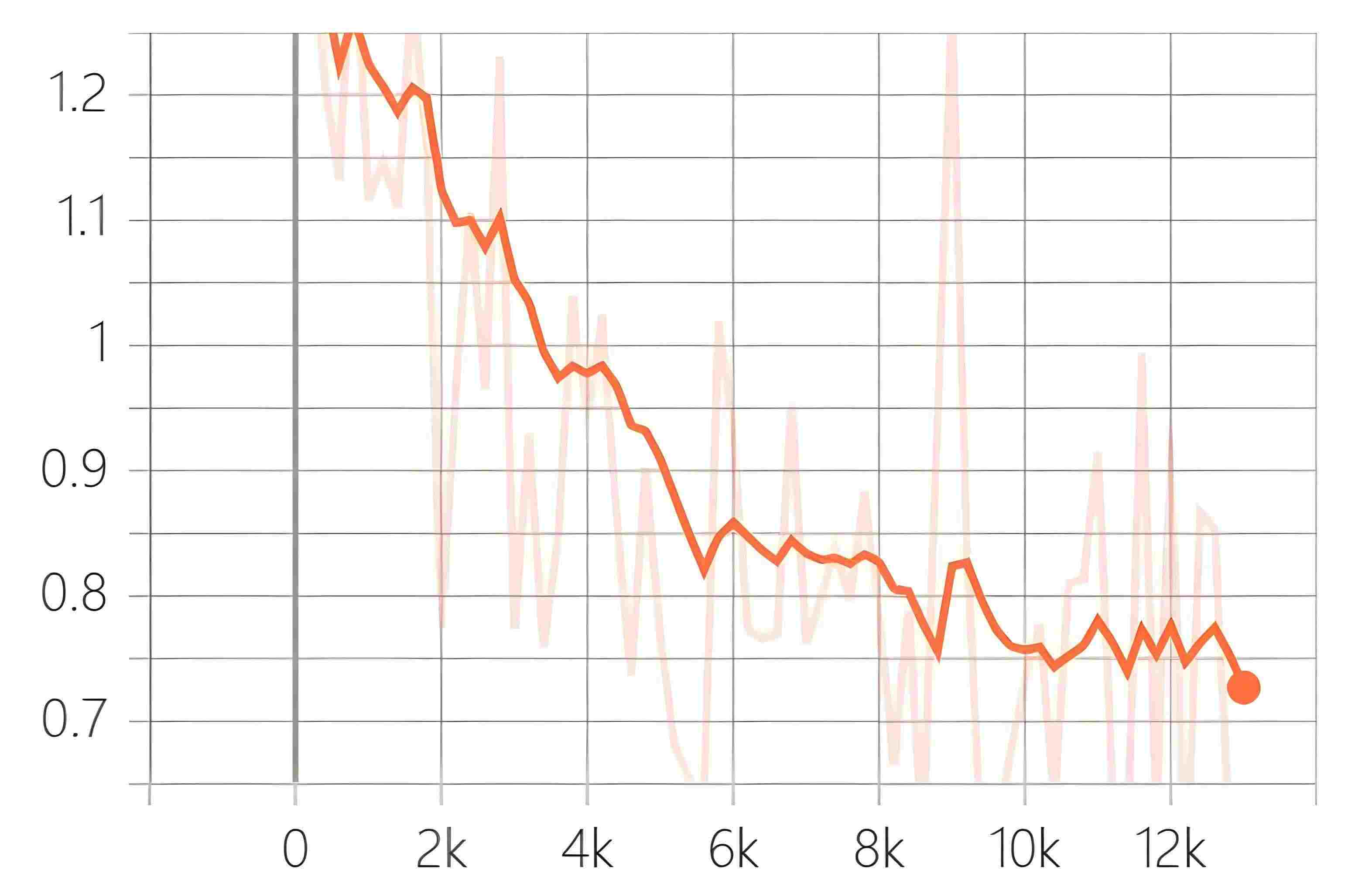}
		\caption{Training trails of NYU v2 2-task}
		\label{fig:nyu2_trails}
	\end{subfigure}
		\hfill
	\begin{subfigure}[b]{0.3\textwidth}
		\centering
		\includegraphics[height=3cm, keepaspectratio]{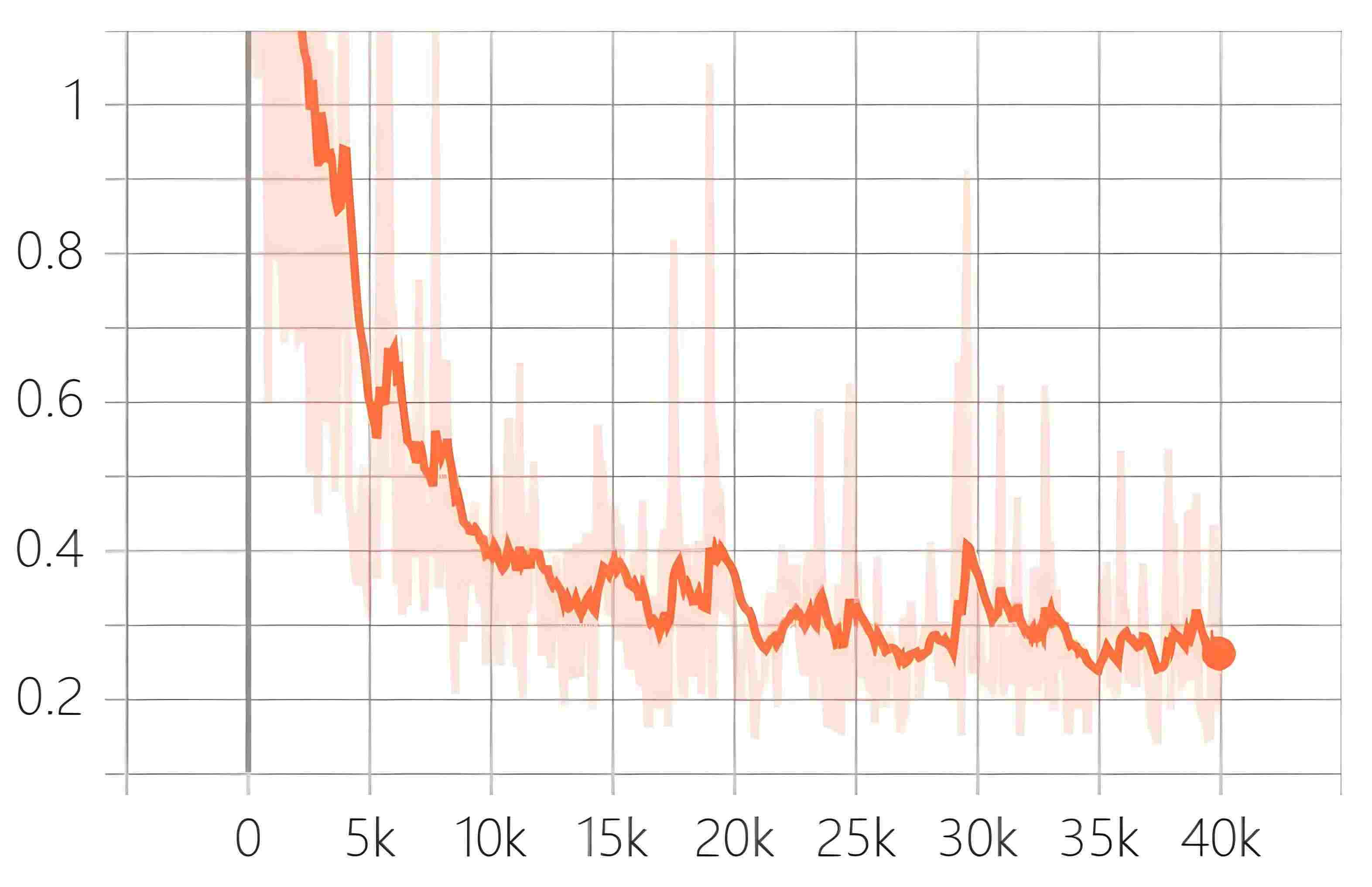}
		\caption{Training trails of Pet 2-task}
		\label{fig:pet_trails}
	\end{subfigure}
	
	\vspace{0.5cm}
		
	\begin{subfigure}[b]{0.3\textwidth}
		\centering
		\includegraphics[height=3cm, keepaspectratio]{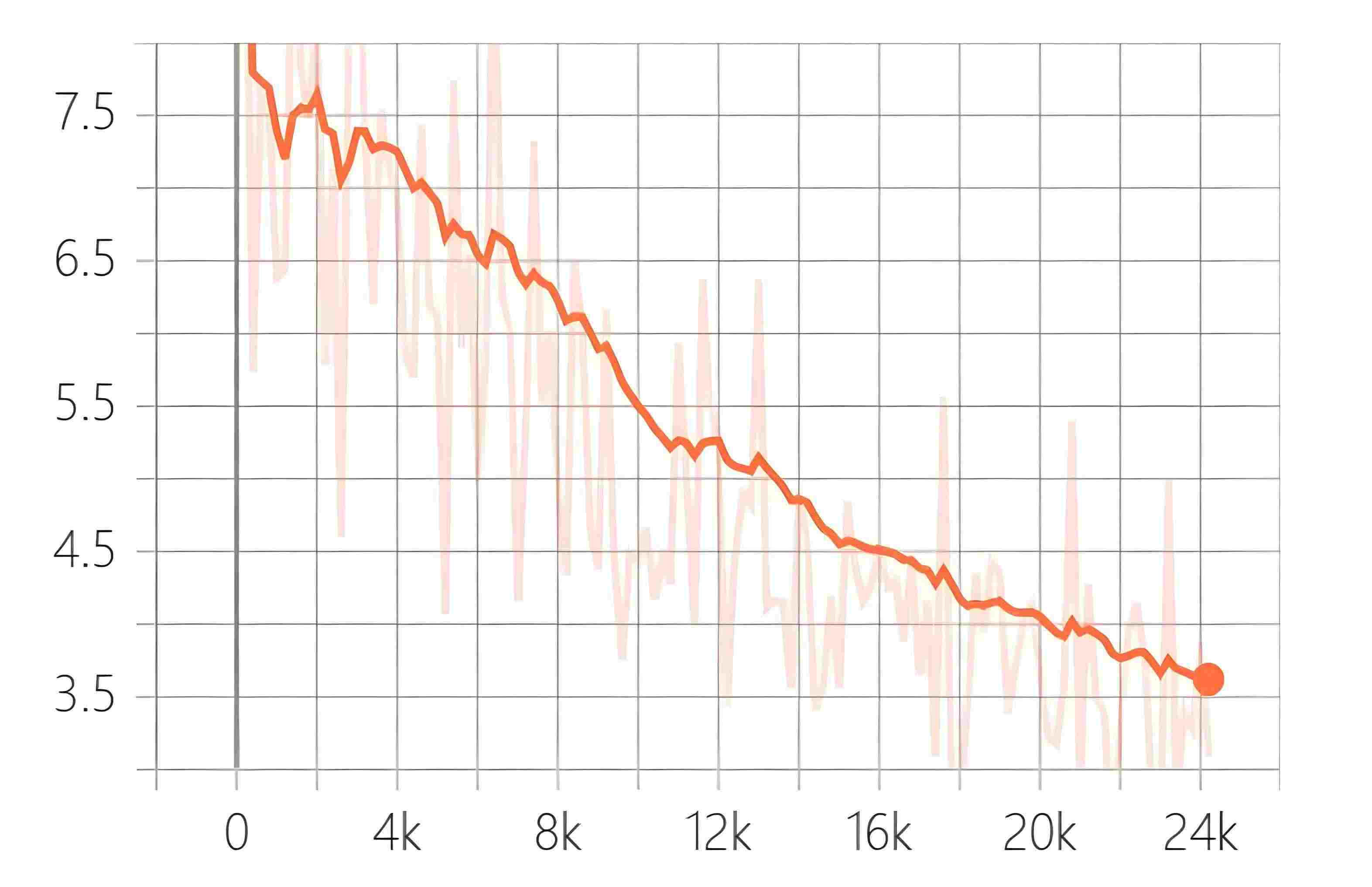}
		\caption{Training trails of NYU v2 3-task}
		\label{fig:nyu3_trails}
	\end{subfigure}
	\hfill
	\begin{subfigure}[b]{0.3\textwidth}
		\centering
		\includegraphics[height=3cm, keepaspectratio]{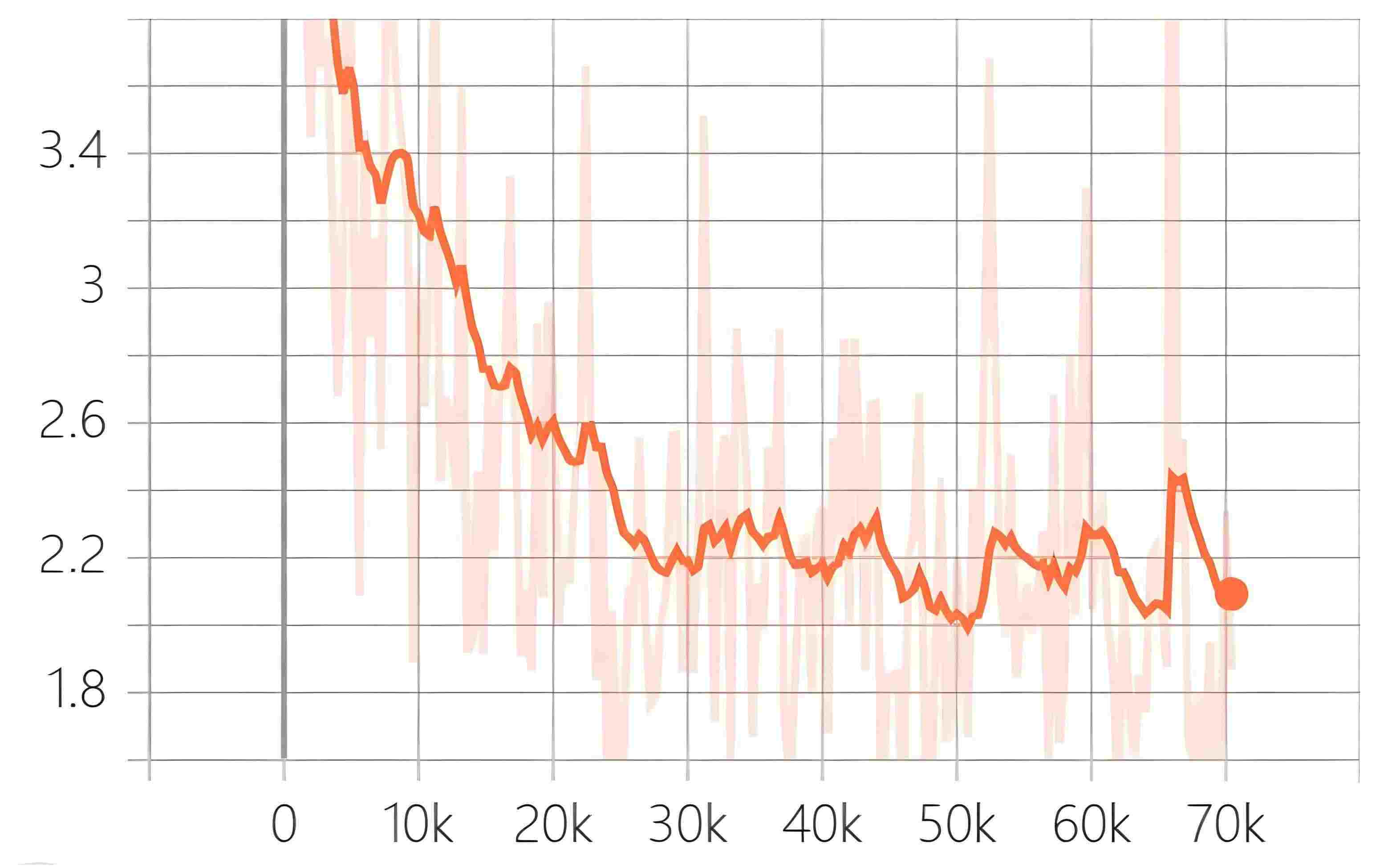}
		\caption{Training trails of Taskonomy 5-task}
		\label{fig:taskonomy_trails}
	\end{subfigure}
	\hfill
	\begin{subfigure}[b]{0.3\textwidth}
		\centering
		\includegraphics[height=3cm, keepaspectratio]{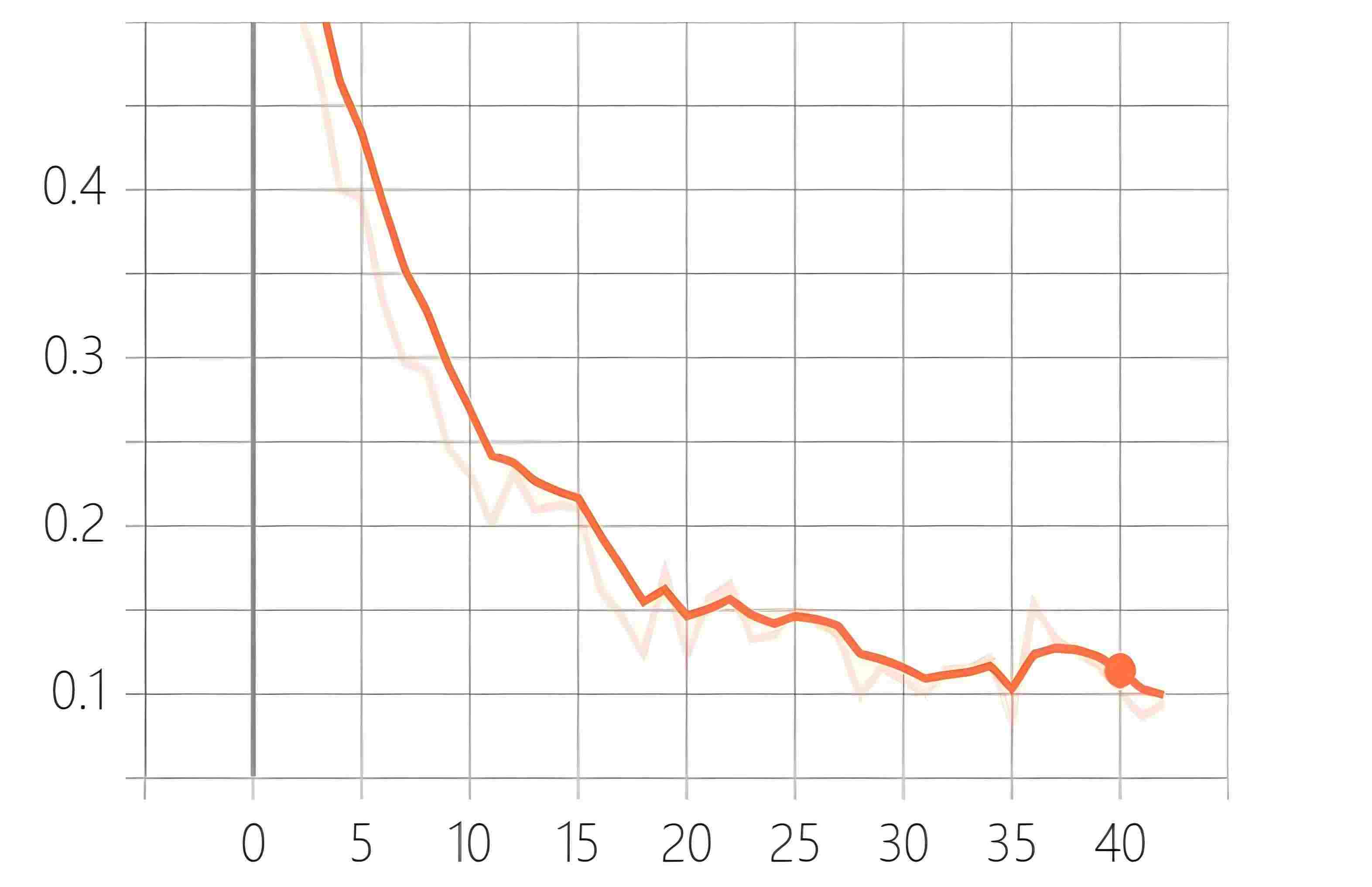}
		\caption{Training trails of MVSA 3-task}
		\label{fig:mvsa_trails}
	\end{subfigure}

	\caption{\small Training trails for different datasets and tasks. The $x$-axis represents the number of training steps, and the $y$-axis represents the training loss.}
	\label{fig:all_trails}
	
\end{figure*}

Additionally, Tables~\ref{tab:cityscapes_results} to \ref{tab:std_results} show the standard deviations (STDs) of the performance metrics, where different random seeds are used to introduce variations in parameter initialization and data shuffling during different runs. The STDs across runs reflect the stability of the models. Fig.~\ref{fig:all_trails} plots the variations of the training  processes. 
Generally, the STDs for the proposed GAI model are relatively small, indicating the stability of the model across different trials.

\begin{table}[t]
\centering
\footnotesize
\caption{Performance and standard deviation on Cityscapes 2-task across different random seeds}
\label{tab:cityscapes_results}
\begin{tabular}{lccccc}
	\toprule
	\textbf{Metric}            & \textbf{Seed 45} & \textbf{Seed 50} & \textbf{Seed 80} & \textbf{Mean} & \textbf{STD}  \\ 
	\midrule
	\textbf{mIoU}              & 46.8             & 46.7             & 46.2             & 46.57         & 0.32          \\ 
	\textbf{Pixel Acc}         & 75.4             & 75.4             & 75.4             & 75.40         & 0.00          \\ 
	\textbf{abs\_err}          & 0.015            & 0.014            & 0.015            & 0.0147        & 0.00058       \\ 
	\textbf{rel\_err}          & 0.300            & 0.309            & 0.300            & 0.3030        & 0.00520       \\ 
	\textbf{$\delta < 1.25$}       & 76.9             & 77.141           & 76.9             & 76.98         & 0.14          \\ 
	\textbf{$\delta < 1.25^2$} & 90.8   & 90.511          & 90.7             & 90.67         & 0.12          \\ 
	\textbf{$\delta < 1.25^3$} & 95.4   & 95.239          & 95.3             & 95.31         & 0.07          \\ 
	\bottomrule
\end{tabular}
\end{table}

\begin{table}[htbp]
\centering
\footnotesize
\caption{Performance and standard deviation on NYUv2 3-task across different random seeds}
\label{tab:nyu_std_results}
\begin{tabular}{lccccc}
	\toprule
	\textbf{Metric} & \textbf{Seed 50} & \textbf{Seed 100} & \textbf{Seed 2048} & \textbf{Mean} & \textbf{STD} \\
	\midrule
	\textbf{mIoU} & 30.200 & 30.100 & 30.300 & 30.2000 & 0.0816 \\
	\textbf{Pixel Accuracy} & 61.400 & 61.600 & 61.800 & 61.6000 & 0.1633 \\
	\textbf{Angular Mean} & 16.200 & 16.641 & 16.399 & 16.4133 & 0.1813 \\
	\textbf{Angular Median} & 13.100 & 12.911 & 12.730 & 12.9137 & 0.1538 \\
	\textbf{Angular 11.25} & 43.700 & 44.965 & 45.449 & 44.7047 & 0.7226 \\
	\textbf{Angular 22.5} & 73.800 & 71.753 & 72.421 & 72.6580 & 1.0342 \\
	\textbf{Angular 30} & 85.300 & 82.956 & 83.609 & 83.9550 & 1.0535 \\
	\textbf{Abs Error} & 0.510 & 0.524 & 0.511 & 0.5150 & 0.0062 \\
	\textbf{Rel Error} & 0.200 & 0.203 & 0.200 & 0.2010 & 0.0014 \\
	\textbf{$\delta < 1.25$} & 67.700 & 65.967 & 67.494 & 67.0537 & 0.9086 \\
	\textbf{$\delta < 1.25^2$} & 91.600 & 90.996 & 91.251 & 91.2823 & 0.2532 \\
	\textbf{$\delta < 1.25^3$} & 97.900 & 97.845 & 97.779 & 97.8413 & 0.0508 \\
	\bottomrule
\end{tabular}
\end{table}

\begin{table}[htbp]
\centering
\footnotesize
\caption{Performance and standard deviation on Tasknonmy 5-task across different random seeds}
\label{tab:std_results}
\begin{tabular}{lccccc}
\toprule
\textbf{Metric}            & \textbf{Seed 45} & \textbf{Seed 50} & \textbf{Seed 80} & \textbf{Mean} & \textbf{STD}  \\ 
\midrule
\textbf{Depth Abs Err}     & 0.0209           & 0.0212           & 0.0210           & 0.0210        & 0.00015       \\ 
\textbf{SN Simi}          & 0.8275           & 0.8324           & 0.8310           & 0.8303        & 0.00253       \\ 
\textbf{Keypoint Err}     & 0.1902           & 0.1905           & 0.1900           & 0.1902        & 0.00025       \\ 
\textbf{Edge Err}          & 0.1963           & 0.1965           & 0.1960           & 0.1963        & 0.00025       \\ 
\textbf{Seg Err}         & 0.4802           & 0.4636           & 0.4620           & 0.4686        & 0.01010       \\ 
\bottomrule
\end{tabular}
\end{table}

\subsection{Computational Overhead}
\label{append:Computational Cost Analysis}
We analyze the computational cost of the proposed GAI model, compared to the existing models, including Cross-Stitch, Sluice, NDDR-CNN, MTAN, DEN, and AdaShare.
The key metrics include computational complexity (measured in GFLOPs) and inference time across multiple tasks.
As shown in Table \ref{tab:inference time comparison}, the GAI model maintains a consistent computational cost of 19.73 GFLOPs, significantly lower than the benchmarks across the considered datasets with different numbers of tasks. 
Note that the GFLOP values in Table~\ref{tab:inference time comparison} only account for the convolutional layers of the models because convolution operations typically dominate the computational cost of deep learning models~\cite{ada}. 

We also measure the inference time of the proposed models on actual hardware. The proposed GAI model achieves a reasonably short inference time across all tasks. For instance, the model requires inference time of 20.9 ms on the CityScapes 2-Task, close to the 19.15 ms inference time of the best-performing benchmark, AdaShare. On the Taskonomy 5-Task, the GAI model requires inference time of 27.58 ms, significantly faster than all benchmarks.

\begin{table}[htbp]
	\caption{Comparison of GFLOPs and Inference Time (ms) for different models across datasets with varying numbers of tasks}
	\centering
	\footnotesize
 \resizebox{\columnwidth}{!}{
	\begin{tabular}{lcccccc}
		\toprule
		\multirow{2}{*}{Models} & \multicolumn{3}{c}{GFLOPs} & \multicolumn{3}{c}{Inference Time (ms)} \\ \cmidrule(lr){2-4} \cmidrule(lr){5-7}
		& \makecell{CityScapes\\2-Task} & \makecell{NYU v2\\3-Task} & \makecell{Taskonomy\\5-Task} & \makecell{CityScapes\\2-Task} & \makecell{NYU v2\\3-Task} & \makecell{Taskonomy\\5-Task} \\ \midrule
		Cross-Stitch            & 37.06              & 55.59         & 92.64            & 21.29              & 32.36         & 57.64            \\ 
		Sluice                  & 37.06              & 55.59         & 92.64            & 21.29              & 32.36         & 57.64            \\ 
		NDDR-CNN                & 38.32              & 57.21         & 100.55           & 20.21              & 30.63         & 52.34            \\ 
		MTAN                    & 44.31              & 58.43         & 82.99            & 29.68              & 40.08         & 60.96            \\ 
		DEN                     & 39.18              & 57.71         & 94.77            & 26.10              & 38.30         & 62.41            \\ 
		AdaShare                & 33.35              & 50.13         & 87.75            & \textbf{19.15}              & 28.96         & 51.01            \\ 
		\textbf{Proposed}   & \textbf{19.73}              & \textbf{19.73}        & \textbf{19.73}            & 20.9               & \textbf{24.5}          & \textbf{27.58}            \\ 
		\bottomrule
	\end{tabular}
 }
	\label{tab:inference time comparison}
\end{table}

\section{Conclusion}

We proposed a novel GAI module to efficiently strengthen the features extracted by the encoder of a multi-task semantic communication system, especially in constrained communication channels. Specifically, we interpreted the outputs of the feature extraction blocks of the encoder into a graph. We also proposed updating the node representation using GAT to capture relationships between the features and using a multi-layer perceptual network to refine the weights between tasks and features. 
Experiments demonstrated the effectiveness of the proposed GAI model in enhancing multi-task semantic communications.
Specifically, GAI surpasses the leading publicly available models by 11.4\% on the CityScapes 2Task dataset and exceeds the current state-of-the-art by 3.97\% on the NYU V2 3Task dataset, respectively, under the constraint of a communication channel bandwidth ratio as low as \(R=\frac{1}{12}\). GAI also demonstrated consistent robustness across a wide SNR spectrum, and the relative accuracy improvement of multi-task semantic communication over its single-task counterpart enlarges with the SNR.

While excelling in optimizing task-specific performance, the GAI model requires retraining for each new task.
We plan to integrate the GAI model with meta-learning techniques, e.g., MAML or Reptile~\cite{nichol2018reptile,finn2017maml}, 
to fine-tune the model when new tasks are added.
Specifically, MAML optimizes initial parameters that can be fine-tuned quickly for a new task, while Reptile can offer a more efficient alternative by approximating MAML through first-order updates. 
Moreover, 
strategies, such as distributed training, model pruning, or more efficient architectures, can be helpful for the scalability of the model. 

\bibliographystyle{IEEEtran}
\bibliography{ref}

   \begin{IEEEbiography}[{\includegraphics[width=1in,height=1.25in,clip,keepaspectratio]{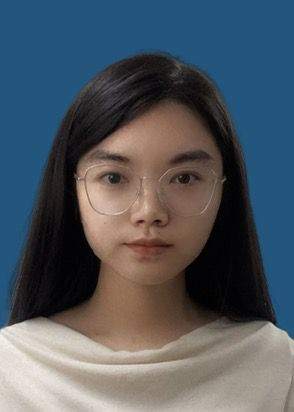}}]{Xi Yu (Graduate Student Member, IEEE)} 
received the B.E. degree in communication engineering from Beijing University of Posts and Telecommunications (BUPT), China, in 2020. She is pursuing her Ph.D. with the School of Information and Communication Engineering at BUPT. Her research interests include multi-task semantic communication and privacy-preserving techniques.
\end{IEEEbiography}\vspace{-20 mm}

\begin{IEEEbiography}[{\includegraphics[width=1in,height=1.25in,clip,keepaspectratio]{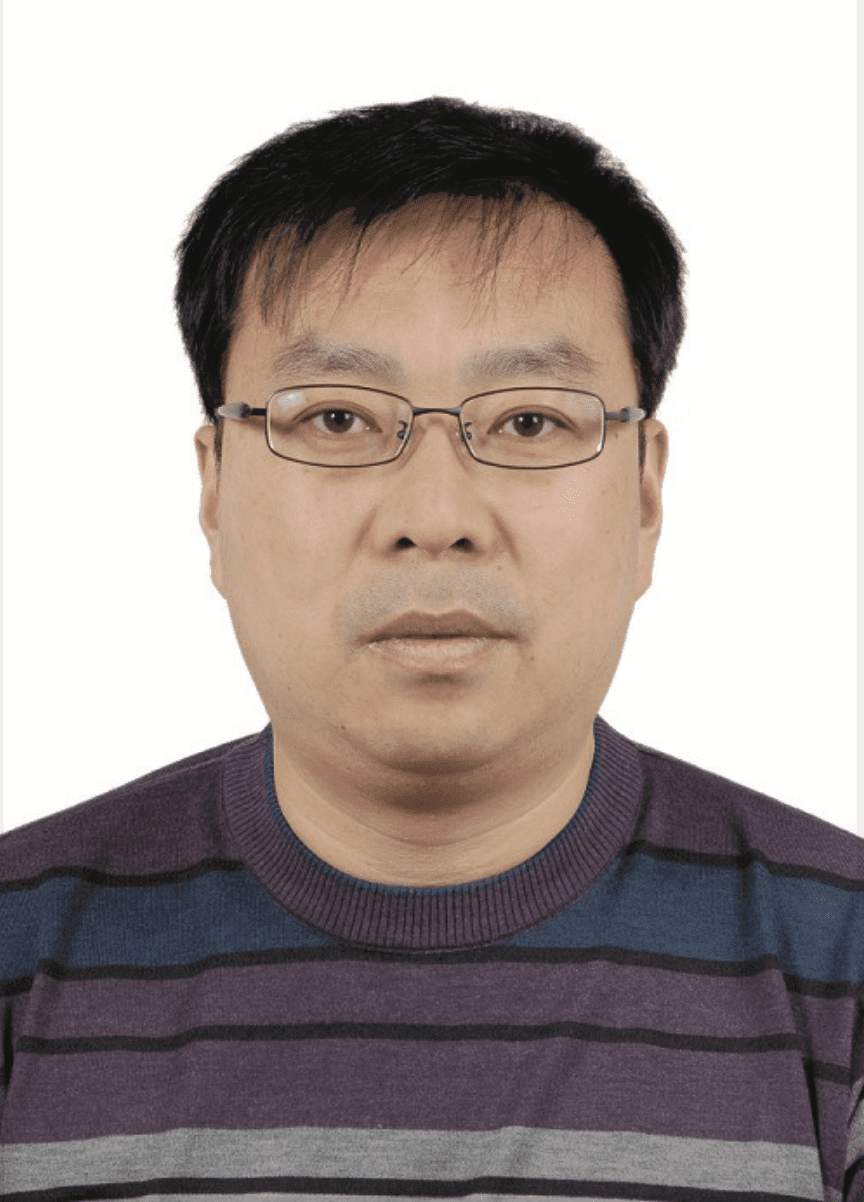}}]{Tiejun Lv (Senior Member, IEEE)} received the M.S. and Ph.D. degrees in electronic engineering from the University of Electronic Science and Technology of China (UESTC), Chengdu, China, in 1997 and 2000, respectively. From January 2001 to January 2003, he was a Postdoctoral Fellow at Tsinghua University, Beijing, China. In 2005, he was promoted to Full Professor at the School of Information and Communication Engineering, Beijing University of Posts and Telecommunications (BUPT). From September 2008 to March 2009, he was a Visiting Professor with the Department of Electrical Engineering at Stanford University, Stanford, CA, USA. He is the author of four books, more than 150 published journal papers and 200 conference papers on the physical layer of wireless mobile communications. His current research interests include signal processing, communications theory and networking. He was the recipient of the Program for New Century Excellent Talents in University Award from the Ministry of Education, China, in 2006. He received the Nature Science Award from the Ministry of Education of China for the hierarchical cooperative communication theory and technologies in 2015. He has won best paper award at CSPS 2022.
\end{IEEEbiography}\vspace{-10 mm}

\begin{IEEEbiography}[{\includegraphics[width=1in,height=1.25in,clip,keepaspectratio]{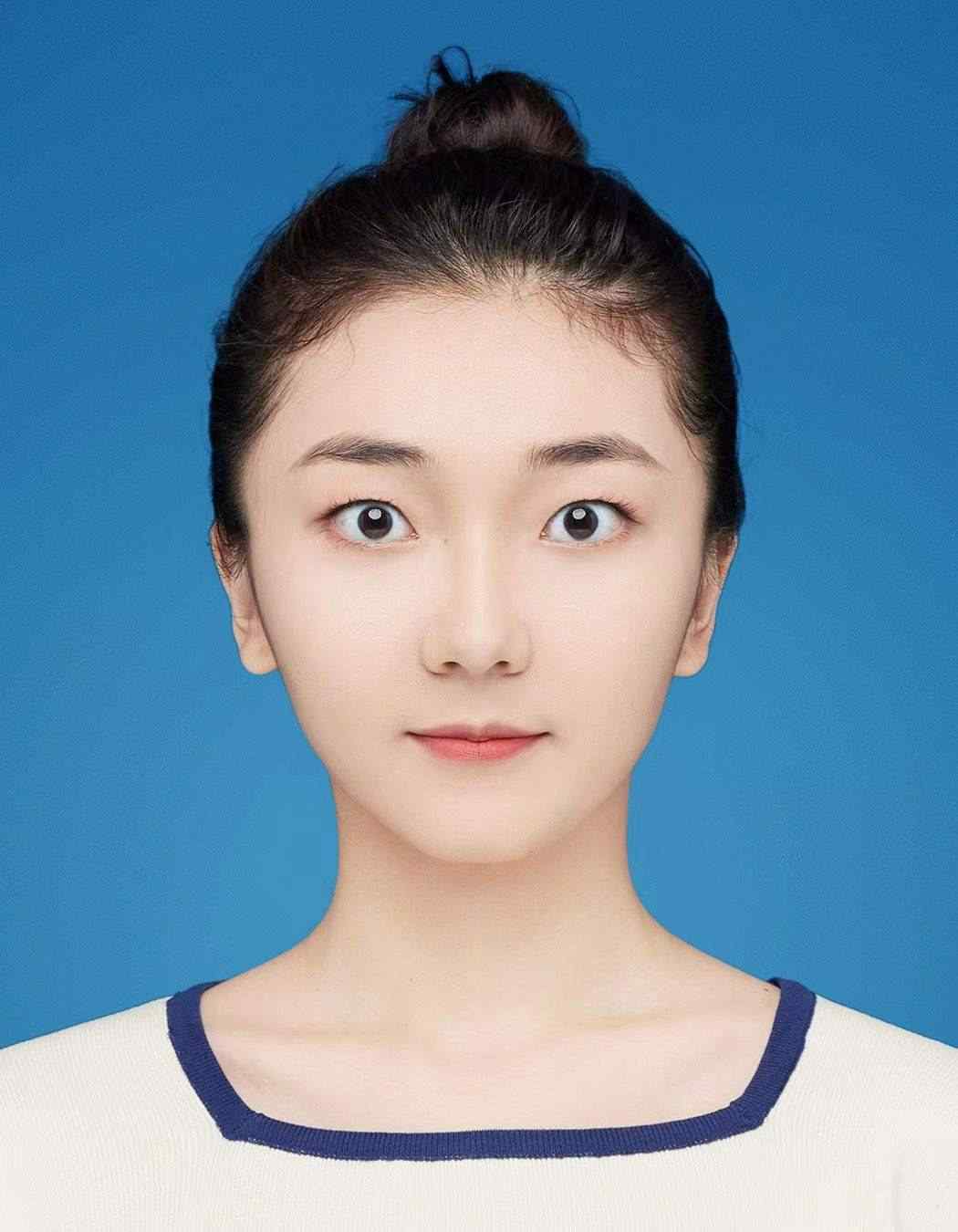}}]{Weicai Li (Graduate Student Member, IEEE)} 
received the B.E. degree in communication engineering from Beijing University of Posts and Telecommunications (BUPT), China, in 2020. She is pursuing her Ph.D. with the School of Information and Communication Engineering at BUPT. From December 2022 to December 2023, she was a Visiting Scholar at the University of Technology Sydney. Her research interests include wireless federated learning, distributed computing, and privacy-preserving.
\end{IEEEbiography}\vspace{-10 mm}

\begin{IEEEbiography}
[{\includegraphics[width=1in,height=1.25in,clip,keepaspectratio]{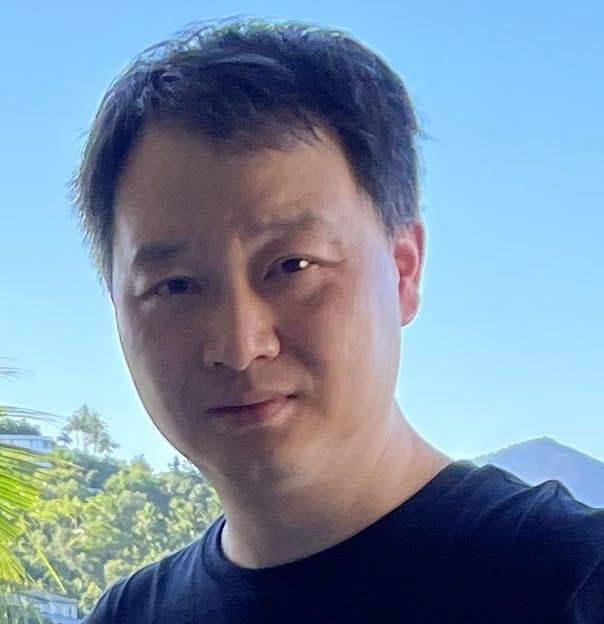}}]
{Wei Ni (F'24)} received his Ph.D. degree in electrical and electronic engineering from Fudan University, Shanghai, China. Currently, he is a Conjoint Research Professor at the University of Technology Sydney, Sydney, New South Wales, Australia. His research interests include communications and signal processing, networking, multiple access, modulation and coding, radio access network (RAN), blockchain, Internet-of-Things (IoT), autonomous systems, and cyber-physical systems (CPS).
\end{IEEEbiography} \vspace{-10 mm}

\begin{IEEEbiography}
[{\includegraphics[width=1in,height=1.25in, clip,keepaspectratio]{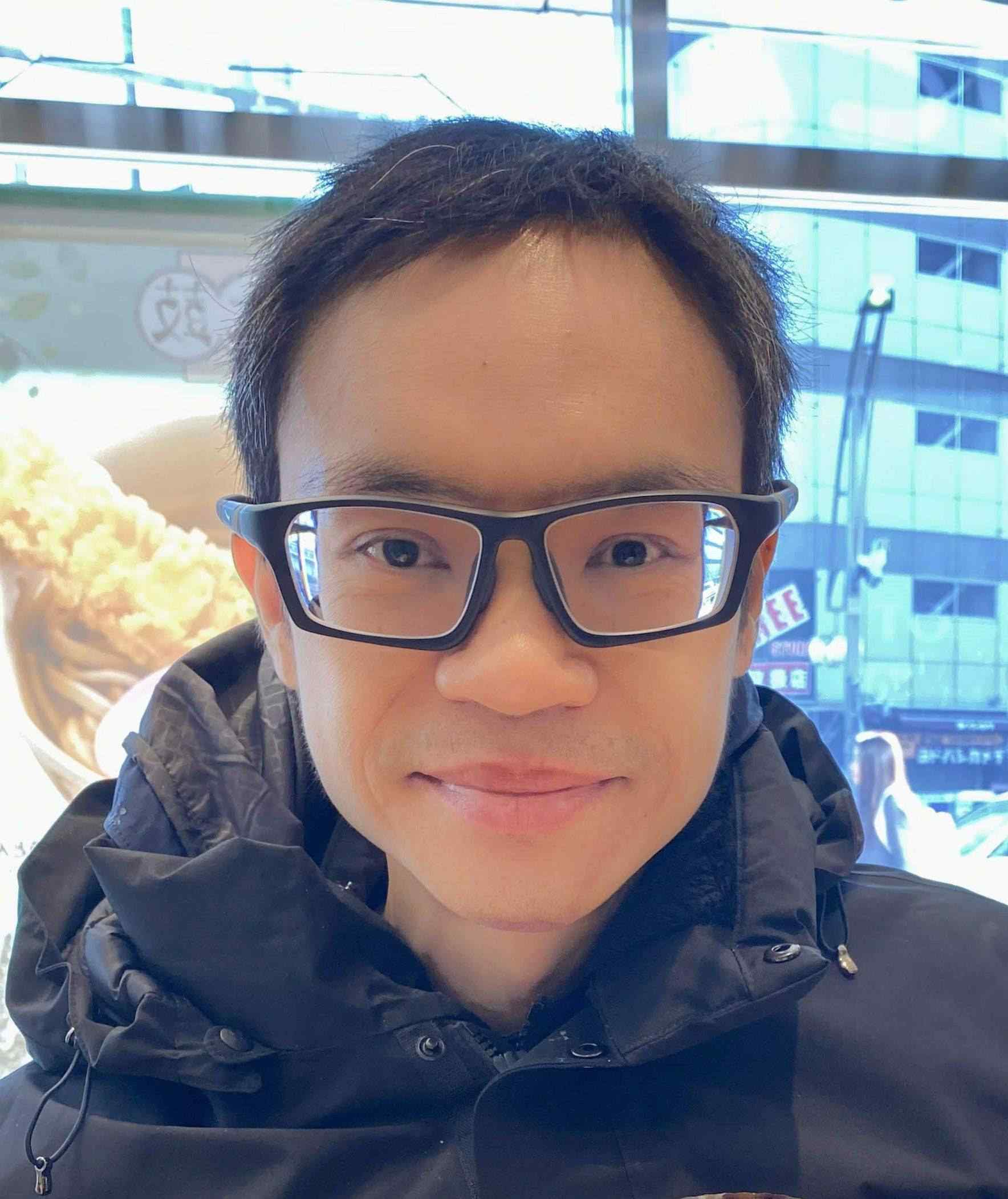}}]{\textbf{Dusit Niyato} (F'17)} 
 received his B.Eng. from the King Mongkut’s Institute of Technology Ladkrabang, Thailand, in 1999, and the Ph.D. degree in electrical and computer engineering from the University of Manitoba, Canada, in 2008. He is currently a Professor of the School of Computer Science and Engineering, Nanyang Technological University, Singapore. His research interests include energy harvesting for wireless communication, Internet of Things, and sensor networks.
\end{IEEEbiography} \vspace{-10 mm}

\begin{IEEEbiography}
[{\includegraphics[width=1in,height=1.25in, clip,keepaspectratio]{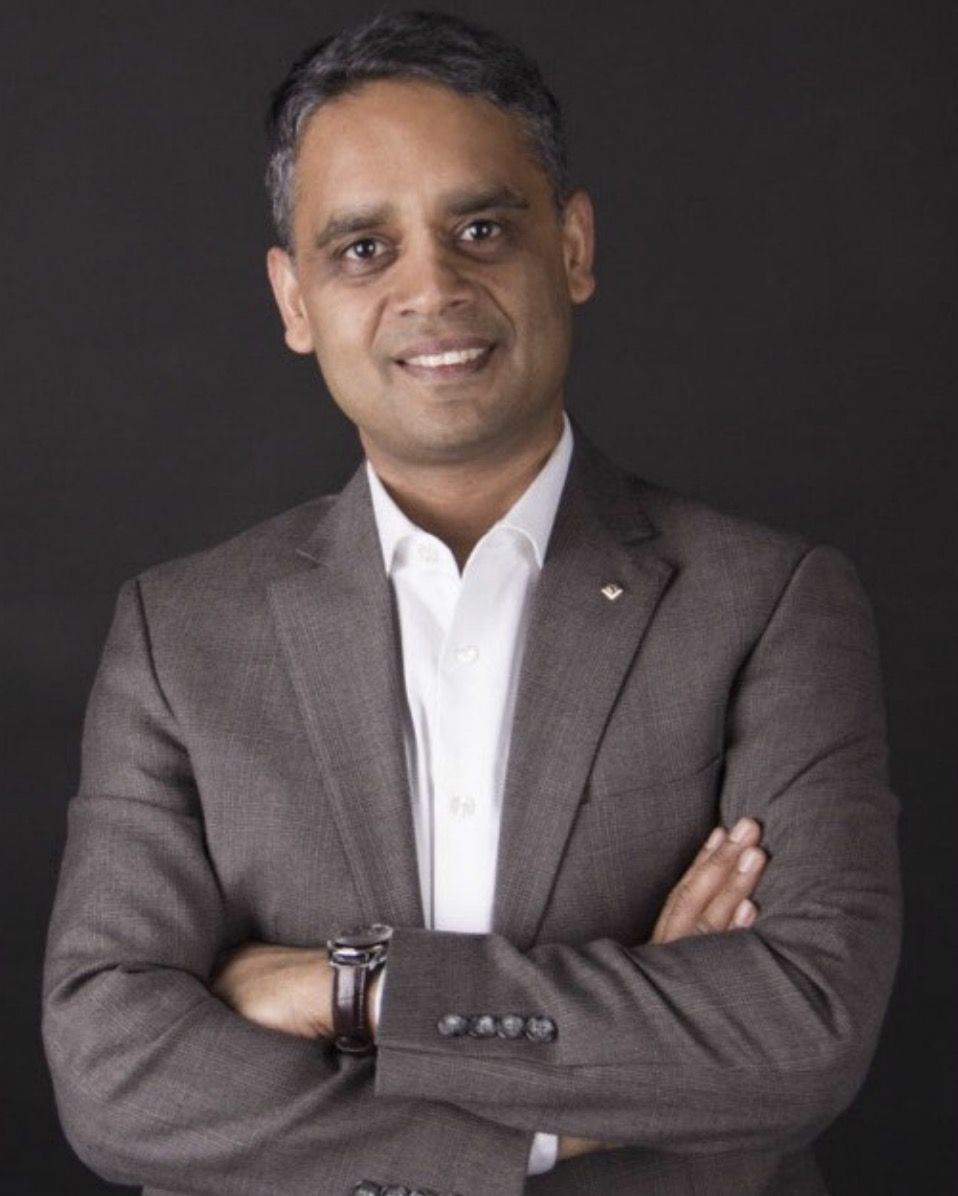}}]{\textbf{Ekram Hossain} (F'15)} 
 is a Professor and the Associate Head (Graduate Studies) of the Department of Electrical and Computer Engineering, University of Manitoba, Canada. He is a Member (Class of 2016) of the College of the Royal Society of Canada. He is also a Fellow of the Canadian Academy of Engineering and the Engineering Institute of Canada. His current research interests include machine learning and AI for optimization of beyond 5G/6G cellular wireless networks.
\end{IEEEbiography}
\end{document}